\documentclass[prl,twocolumn,twoside,byrevtex,superscriptaddress,showpacs,floatfix]{revtex4-1}

\usepackage{currfile}
\usepackage{graphicx,epsfig,verbatim,enumerate}
\usepackage{amssymb,amsmath}
\usepackage{ifthen,tikz}

\usepackage{hyperref}
\hypersetup{
        colorlinks = true,
        allcolors = blue
}

\usepackage{longtable}

\usepackage{mathtools}

\newboolean{twocolswitch}

\newcommand{\sindex}[1]{}
\newcommand{\nindex}[1]{}

\newcommand{\www}[1]{\url{#1}}

\usepackage{lettrine}

\usepackage{color}

\newcommand{\partitionprob}{q}
\newcommand{\partitiontype}[1]{$\partitionprob$$=$$#1$}

\newcommand{\onehalf}{\frac{1}{2}}
\newcommand{\onequarter}{\frac{1}{4}}

\newcommand{\xmin}{r_{\textnormal{max}}}
\newcommand{\kstat}{D}
\newcommand{\clp}{\mbox{$p$-value}}

\setboolean{twocolswitch}{true}

\begin{document}

\title{\protect
Zipf's law holds for phrases, not words
}

\author{
\firstname{Jake Ryland}
\surname{Williams}
}

\email{jake.williams@uvm.edu}

\affiliation{Department of Mathematics \& Statistics,
  Vermont Complex Systems Center,
  Computational Story Lab,
  \& the Vermont Advanced Computing Core,
  The University of Vermont,
  Burlington, VT 05401.}

\author{
\firstname{Paul R.}
\surname{Lessard}
}

\email{paul.lessard@boulder.edu}

\affiliation{Department of Mathematics,
  University of Colorado,
  Boulder
  CO, 80309}

\author{
\firstname{Suma}
\surname{Desu}
}

\email{sdesu@mit.edu}

\affiliation{
  Center for Computational Engineering,
  Massachusetts Institute of Technology,
  Cambridge,
  MA, 02139
}

\author{
\firstname{Eric M.}
\surname{Clark}
}

\email{eric.clark@uvm.edu}

\affiliation{Department of Mathematics \& Statistics,
  Vermont Complex Systems Center,
  Computational Story Lab,
  \& the Vermont Advanced Computing Core,
  The University of Vermont,
  Burlington, VT 05401.}

\author{
\firstname{James P.}
\surname{Bagrow}
}

\email{james.bagrow@uvm.edu}

\affiliation{
  Computational Story Lab,
  Vermont Advanced Computing Core,
  \& the Department of Mathematics and Statistics, University of Vermont,
  Burlington,
  VT, 05401
}
\affiliation{
  Vermont Complex Systems Center,
  University of Vermont,
  Burlington,
  VT, 05401
}

\author{
\firstname{Christopher M.}
\surname{Danforth}
}
\email{chris.danforth@uvm.edu}

\affiliation{Department of Mathematics \& Statistics,
  Vermont Complex Systems Center,
  Computational Story Lab,
  \& the Vermont Advanced Computing Core,
  The University of Vermont,
  Burlington, VT 05401.}

\author{
\firstname{Peter Sheridan}
\surname{Dodds}
}
\email{peter.dodds@uvm.edu}

\affiliation{Department of Mathematics \& Statistics,
  Vermont Complex Systems Center,
  Computational Story Lab,
  \& the Vermont Advanced Computing Core,
  The University of Vermont,
  Burlington, VT 05401.}

\date{\today}

\begin{abstract}
  \protect
  With Zipf's law being originally and 
most famously observed for word frequency, 
it is surprisingly limited in its applicability to human language,
holding over no more than three to four orders of magnitude before hitting
a clear break in scaling.
Here, building on the simple observation that phrases of one or more words
comprise the most coherent units of meaning in language,
we show empirically that Zipf's law for phrases
extends over as many as nine orders of rank magnitude.
In doing so, 
we develop a principled and scalable statistical mechanical
method of random text partitioning,
which opens up a rich frontier of rigorous text 
analysis via a rank ordering of mixed length phrases.
 
\end{abstract}


\maketitle

\section{Introduction}

Over the last century, the elements of many disparate systems 
have been found to approximately follow 
Zipf's law---that element size is inversely proportional
to element size rank~\cite{zipf1935a,zipf1949a}---from 
city populations~\cite{zipf1949a,simon1955a,batty2008a},
to  firm sizes~\cite{axtell2001a},
and
family names~\cite{zanette2001a}.
Starting with Mandelbrot's optimality argument~\cite{mandelbrot1953a},
and the dynamically growing, rich-get-richer model of
Simon~\cite{simon1955a}, strident debates over theoretical
mechanisms leading to Zipf's law have continued until the
present~\cite{miller1957a,ferrericancho2010a,dsouza2007a,coromina-murtra2010a}.
Persistent claims of uninteresting randomness underlying Zipf's law~\cite{miller1957a}
have been successfully challenged~\cite{ferrericancho2010a}, and
in non-linguistic systems, good evidence supports Simon's
model~\cite{simon1955a,bornholdt2001a,maillart2008a}
which has been found to be the basis of scale-free
networks~\cite{price1976a,barabasi1999a}.

For language, the vast majority of arguments have focused on the frequency of an individual word
which we suggest here is the wrong fundamental unit of analysis.
Words are an evident building block of language, and we are naturally
drawn to simple counting as a primary means of analysis
(the earliest examples are Biblical corcordances, dating to the 13th Century).
And while we have defined morphemes as the most basic meaningful `atoms' of language,
the meaningful `molecules' of language are 
clearly a mixture of individual words and phrases.
The identification of meaningful phrases, or multi-word expressions, 
in natural language poses one of the largest obstacles to accurate machine
translation~\cite{sag2002a}.
In reading the phrases ``New York City'' or ``Star Wars'',
we effortlessly take them as irreducible constructions, different
from the transparent sum of their parts.  
Indeed, it is only with some difficulty that we actively parse 
highly common phrases and consider their individuals words.

While partitioning a text into words is straightforward
computationally, 
partitioning into meaningful phrases would appear to require a next
level of sophistication requiring online human analysis.
But in order to contend with the increasingly very large sizes and
rapid delivery rates
of important text corpora---such as news and social media---we are obliged to find a simple, 
necessarily linguistically naive, yet effective method.

A natural possibility is to in some way capitalize on $N$-grams, 
which are a now common and fast approach for parsing a text.
Large scale $N$-gram data sets have been made widely available for analysis,
most notably through the Google Books project~\cite{googlebooks-ngrams2014a}.
Unfortunately, all $N$-grams fail on a crucial front:
in their counting they overlap, which obscures underlying word frequencies.
Consequently, and crucially,
we are unable to properly assign 
rankable frequency of usage weights to
$N$-grams combined across all values of $N$.

Here, we introduce `random partitioning',
a method that is fast, intelligible, scalable, 
and sensibly preserves word frequencies: 
i.e., the sum of sensibly-weighted
partitioned phrases is equal to the total number of words present.
As we show, our method immediately yields
the profound basic science result that phrases of mixed lengths, 
as opposed to just individual words, obey Zipf's law,
indicating the method 
can serve as a profitable approach to general text analysis.
To explore a lower level of language, 
we also partition for
sub-word units, or graphemes,
by breaking words into letter sequences.
In the remainder of the paper, we first describe random partitioning 
and then present results for a range of texts.

\section{Text partitioning}

To begin our random partitioning process, 
we break a given text $T$ into clauses,
as demarcated by standard punctuation
(other defensible schemes for obtaining clauses may also be used),
and define the length norm, $\ell$, of a given clause $t$ 
(or phrase, $s\in S$)
as its word count, written $\ell(t)$.
We then define a partition, $\mathcal{P}$, of a clause $t$
to be a sequence of the boundaries surrounding its words:
\begin{equation}
        \mathcal{P}:~x_0<\cdots<x_{\ell(t)},
\end{equation}
and note that $x_0,x_{\ell(t)}\in \mathcal{P}$ for any $\mathcal{P}$,
as we have (a priori) the demarcation knowledge of the clause.
For example, consider the highly ambiguous text:
\begin{center}
``Hot dog doctor!''
\end{center}
Forgoing punctuation and casing,
we might attempt to break the clause down,
and interpret through the partition:
\begin{center}
  \begin{tikzpicture}[%
      back line/.style={densely dotted},
      cross line/.style={preaction={draw=white, -,line width=6pt}}]
        \node at (-1.5,0) (t) {hot};
    \node at (0,0) (t) {dog};
    \node at (1.5,0) (t) {doctor};
        \node at (-4,-0.4) (linl) {};
    \node at (4,-0.4) (linr) {};
        \node at (-3,-1) (P) {$\mathcal{P}:$};
        \node at (-2.25,-1) (x0) {$x_0$};
    \node at (-2.25,-0.8) (b0l) {};
    \node at (-2.25,0.3) (b0h) {};
        \node at (-0.75,-1) (x1) {$x_1$};
    \node at (-0.75,-0.8) (b1l) {};
    \node at (-0.75,0.3) (b1h) {};
        \node at (2.25,-1) (x3) {$x_3$};
    \node at (2.25,-0.8) (b3l) {};
    \node at (2.25,0.3) (b3h) {};
        \draw[cross line,<->, black, line width=1] (linl) -- (linr);
    \draw[cross line,-, gray, line width=1] (b0l) -- (b0h);
    \draw[cross line,-, red, line width=1] (b1l) -- (b1h);
    \draw[cross line,-, gray, line width=1] (b3l) -- (b3h);    
  \end{tikzpicture}
\end{center}
i.e., $\mathcal{P}=\{x_0,x_1,x_3\}$,
which breaks the text into phrases,
``hot'' and ``dog doctor'',
and assume it as reference to an 
attractive veterinarian
(as was meant in \cite{cougarTown2013s4e4}).
However, depending on our choice,
we might have found an alternative meaning:
\begin{equation*}
\begin{split}
\text{hot dog; doctor}&\text{: A daring show-off doctor.}\\
&\text{: One offers a frankfurter to a doctor.}\\
\text{hot; dog doctor}&\text{: An attractive veterinarian (vet).}\\
&\text{: An overheated vet.}\\
\text{hot dog doctor}&\text{: A frank-improving condiment.}\\
&\text{: A frank-improving chef.}\\
\text{hot; dog; doctor}&\text{: An attractive vet of canines.}\\
&\text{: An overheated vet of canines.}
\end{split}
\end{equation*}
Note in the above that we 
(as well as the speaker in \cite{cougarTown2013s4e4}) 
have allowed the phrase ``dog doctor''
to carry idiomatic meaning
in its non-restriction to canines,
despite the usage of the word ``dog''.

Now, in an ideal scenario we might have some knowledge
of the likelihood for each boundary to be ``cut''
(which would produce an `informed' partition method),
but for now our goal is generality, 
and so we proceed, assuming a uniform
boundary-cutting probability, $q$, 
across all $\ell(t)-1$ word-word (clause-internal) boundaries of a clause, $t$.
In general, there are $2^{\ell(t)-1}$ possible partitions of $t$ 
involving $\frac{1}{2} \ell(t) (\ell(t) + 1)$ potential phrases.
For each integral pair 
$i,j$ with $1\leq i<j\leq\ell(t)$, 
we note that the 
probability for a randomly chosen partition of the clause $t$
to include the (contiguous) phrase, $t_{i\cdots j}$,
is determined by successful
cutting at the ends 
of $t_{i\cdots j}$ and failures within 
(e.g., $x_2$ must \emph{not} be cut to produce ``dog~doctor''),
accommodating for $t_{i\cdots j}$ reaching one or both ends of $t$,
i.e.,
\begin{equation}
  \label{eq:tpz.phraseweight}
  P_{\partitionprob}(t_{i\cdots j} \,|\, t) = \partitionprob^{2-b_{i\cdots j}}(1-\partitionprob)^{\ell(s)-1}
\end{equation}
where $b_{i\cdots j}$ is the number of the clause's boundaries shared by $t_{i\cdots j}$ and $t$.
Allowing for a phrase $s\in S$ to have labeling equivalence
to multiple contiguous regions 
(i.e., $s=t_{i\cdots j}=t_{i'\cdots j'}$, with $i,j\neq i',j'$) 
within a clause
e.g., ``ha ha'' within ``ha ha ha'',
we interpret the `expected frequency' of $s$ given the text
by the double sum:
\begin{equation}
  \label{eq:tpz.phraseweight_clause}
  f_q(s \,|\, T)
  =
  \sum_{t \in T} 
  f_q(s \,|\, t)
  =
  \sum_{t \in T} 
  \sum_{s=t_{i\cdots j}}        
  P_{\partitionprob}(t_{i\cdots j} \,|\, t).
\end{equation}
Departing from normal word counts, we may now
have $f_q \ll 1$,
except when one partitions for
word ($q=1$) or clause ($q=0$) frequencies.
When weighted by phrase length,
the partition frequencies of phrases 
from a clause
sum to the total number of words 
originally present in the clause:
\begin{equation}
  \label{eq:tpz.phraseweight_text}
  \ell(t)
  =
  \sum_{1\leq i<j\leq\ell(t)}
  \ell(t_{i\cdots j})P_q(t_{i\cdots j} \,|\, t),
\end{equation}
which ensures that when the expected
frequencies of phrases, $s$, are summed 
(with the length norm) 
over the whole text:
\begin{equation}
  \label{eq:tpz.wordmass_text}
  \sum_{s}
  \ell(s)
  f_q(s\,|\, T)
  =
  \sum_{t\in T}
  \ell(t)f(t),
\end{equation}
the underlying mass of words in the text is conserved
(see SI-2 for proofs of Eqs.~\ref{eq:tpz.phraseweight_text}~and~\ref{eq:tpz.wordmass_text}).
Said differently, 
phrase partition frequencies 
(random or otherwise)
conserve word frequencies
through the length norm $\ell$,
and so have a physically meaningful
relationship to the
words on ``the page.''

\section{Statistical Mechanical  interpretation}

Here, we focus on three natural kinds of partitions:
\partitiontype{0}: clauses are partitioned only as clauses themselves;
\partitiontype{\onehalf}: what we call `pure random partitioning'---all partitions of a clause are equally likely;
\partitiontype{1}: clauses are partitioned into words.

In carrying out pure random partitioning (\partitiontype{\onehalf}),
which we will show has the many desirable properties we seek,
we are assuming all partitions are equally likely,
reminiscent of equipartitioning used in statistical
mechanics~\cite{goldenfeld1992a}.
Extending the analogy, we can view \partitiontype{0} as a zero
temperature limit, and \partitiontype{1} as an infinite
temperature one.
As an anchor for $f_{\onehalf}$, we note that 
words that appear once within a text---hapax legomena---will have 
$f_q \in \{\onequarter,\onehalf, 1\}$
(depending on clause boundaries),
on the order of 1 as per standard word partitioning.

\section{Experiments and Results}

Before we apply the random partition theory
to produce our generalization of word count, $f_q$,
we will first examine the results of applying
the random partition process in a `one-off' manner.
We process through the clauses of a text once,
cutting word-word boundaries
(and in a parallel experiment for graphemes, 
cutting letter-letter boundaries within words)
uniformly at random with probability $q=\frac{1}{2}$.

\begin{figure}[tp!]
  \centering
  \includegraphics[width=0.49\textwidth]{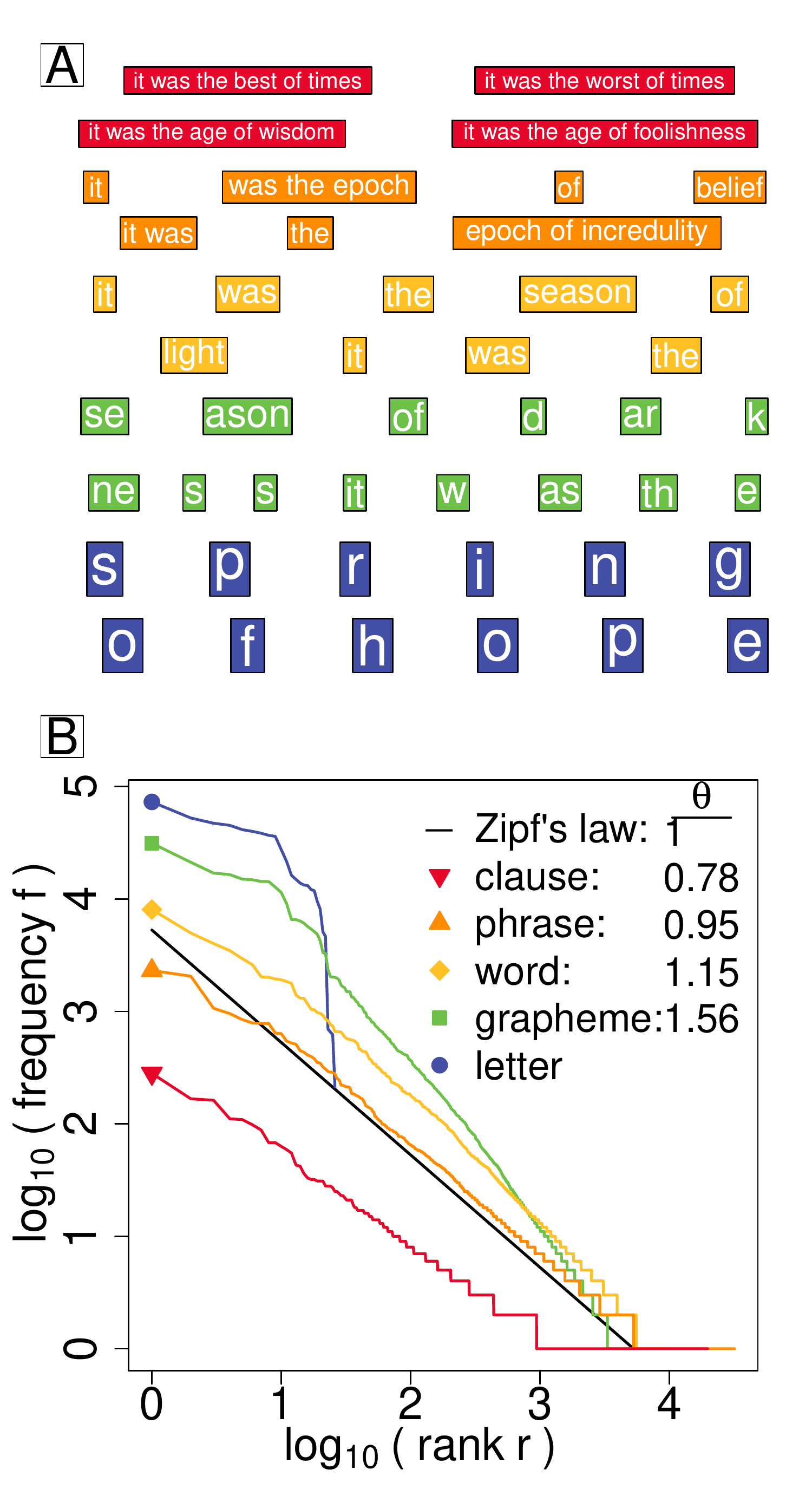}
  \caption{
    \textbf{A.}
    Partition examples for the start of Charles Dickens's ``Tale of Two Cities''
    at five distinct levels:
    clauses (red),
    pure random partitioning phrases ($\partitionprob=\frac{1}{2}$, orange),
    words (yellow),
    pure random partitioning graphemes ($\partitionprob=\frac{1}{2}$,
    green),
    and
    letters (blue).
    The specific phrases and graphemes shown are for one realization
    of pure random partitioning.
    \textbf{B.}
    Zipf distributions for the five kinds of partitions
    along with estimates of the Zipf exponent $\theta$ when
    scaling is observed.
    No robust scaling is observed at the letter scale.
    The colors match those used in panel \textbf{A}, and the symbols
    at the start of each distribution are intended to 
    strengthen the connection to the legend.
    See Ref.~\cite{clauset2009b} and supplementary material for measurement details.
   }
  \label{fig:tpz.twocities_cartoon_zipf}
\end{figure}

In Fig.~\ref{fig:tpz.twocities_cartoon_zipf}A,
we present an example `one-off' partition
of the first few lines of Charles Dickens' ``Tale of Two Cities''
We give example partitions at the scales of
clauses (red),
pure random partition phrases (orange),
words (yellow),
pure random partition graphemes (green),
and 
letters (blue).
In Fig.~\ref{fig:tpz.twocities_cartoon_zipf}B,
we show Zipf distributions for all five partitioning scales.
We see that clauses (\partitiontype{0})
and pure random partitioning phrases
(\partitiontype{\onehalf})
both adhere well to the pure form of 
$f \propto r^{-\theta}$ where $r$ is rank.
For clauses  we find $\theta \simeq 0.78$
and for random partitioning, $\theta \simeq 0.98$
(see supplementary material for measurement details and for examples
of other works of literature).
The quality of scaling degrades as we move down to words and graphemes
with the appearance of scaling breaks~\cite{ferrericancho2001c,gerlach2013a,williams2014b}.
Scaling vanishes entirely at the level of letters.

\begin{figure}[tp!]
  \centering
  \includegraphics[width=0.49\textwidth]{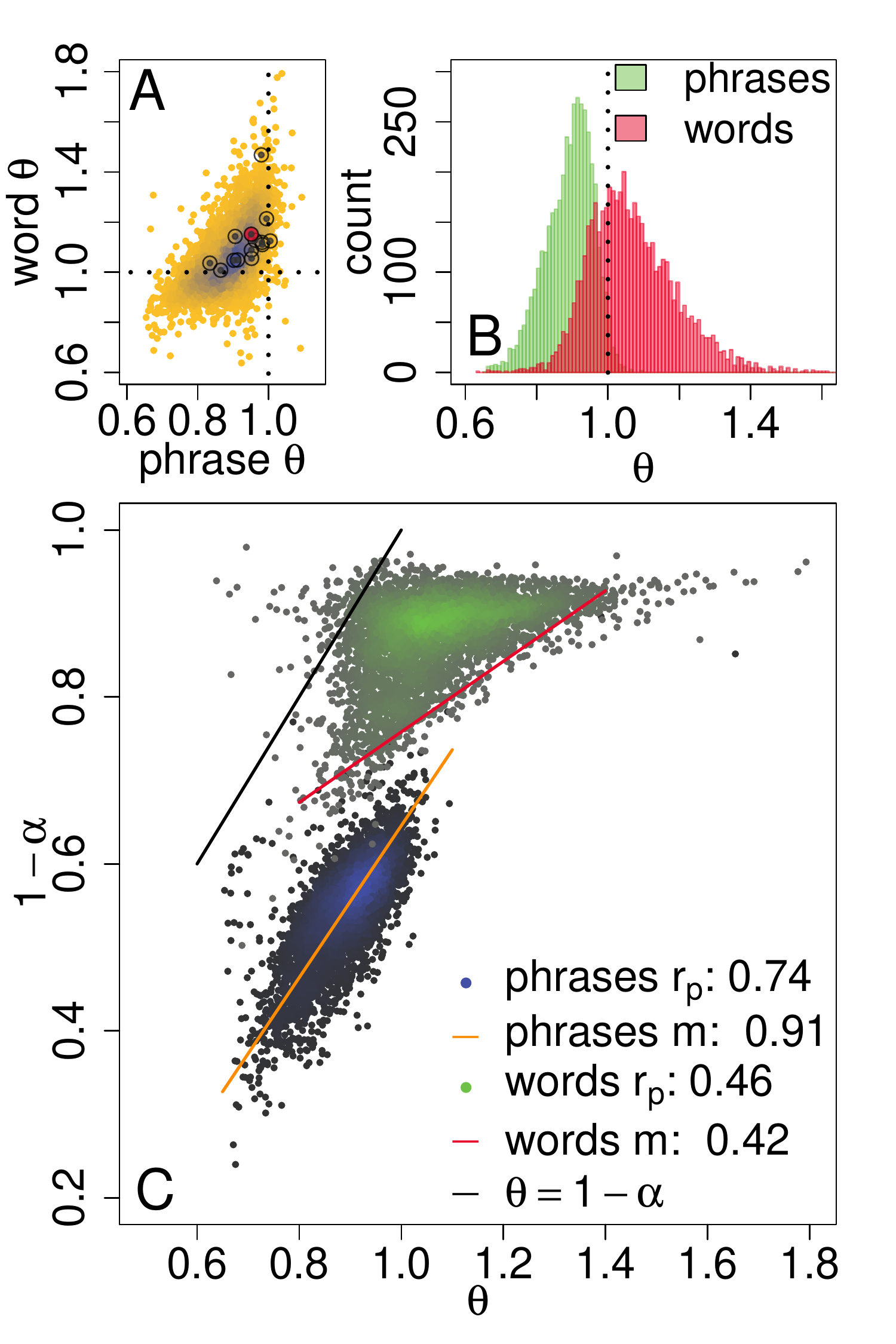}
  \caption{
    \textbf{A.}
    Density plot showing the Zipf exponent $\theta$ for `one-off' 
    randomly
    paritioned phrases and
    word Zipf distributions
    (\partitiontype{1} and \partitiontype{\onehalf})
    for around 4000 works of literature.
    We indicate ``Tale of Two Cities'' by the red circle,
    and with black circles, we represent measurements for 14 other  
    works of literature analyzed further in the supplementary material.
    \textbf{B.}
    Histograms of the Zipf exponent $\theta$ for the same set of books (marginal
    distributions for \textbf{A}).
    Phrases typically exhibit $\theta \le 1$ whereas
    words produce unphysical $\theta > 1$, according to Simon's model
    \textbf{C.}
    Test of Simon's model's analytical connection
    $\theta = 1-\alpha$,
    where 
    $\theta$ is the Zipf exponent
    and 
    $\alpha$ is the rate at 
    which new terms (e.g., graphemes, words, phrases)
    are introduced throughout a text.
    We estimate $\alpha$ as the number of
    different words normalized by the total word volume.
    For both words and phrases, we compute linear fits 
    using Reduced Major Axis (RMA) regression~\cite{rayner1985a}
    to obtain slope $m$, along with
    the Pearson correlation coefficient $r_{\textnormal{p}}$.
    Words (green) do not exhibit a simple linear
    relationship whereas phrases do (blue),
    albeit clearly below
    the $\alpha = 1-\theta$ line in black.
   }
  \label{fig:tpz.twocities_more}
\end{figure}

Moving beyond a single work,
we next summarize findings for 
a large collection of texts~\cite{gutenberg2010}
in Fig.~\ref{fig:tpz.twocities_more}A, 
and compare the Zipf exponent
$\theta$ for words and pure random \partitiontype{\onehalf} 
`one-off' partitioning
for around 4000 works of literature.  
We plot the corresponding 
marginal distributions in Fig.~\ref{fig:tpz.twocities_more}B,
and see that clearly $\theta \lesssim 1$ for 
\partitiontype{\onehalf} phrases, 
while for words, there is a strong positive skew 
with the majority of values of $\theta > 1$.
These steep scalings for words (and graphemes), $\theta > 1$,
are not dynamically accessible for Simon's model~\cite{dsouza2007a}.

Leaving aside this non-physicality 
of Zipf distributions for words
and concerns about breaks in scaling,
we recall that Simon's model connects the rate, $\alpha$,
at which new terms are introduced,
to $\theta$ in a simple way: $1-\alpha=\theta$~\cite{simon1955a}.
Given frequency data from a pure Simon model,
the word/phrase introduction rate is determined
easily to be $\alpha = N/M$, 
where $N$ is the number of unique words/phrases, 
and $M$ is the sum total of all word/phrase frequencies.
We ask how well works of literature conform to this connection
in Fig.~\ref{fig:tpz.twocities_more}C,
and find that  words (green dots)
do not demonstrate any semblance 
of a linear relationship,
whereas phrases (blue dots) exhibit a clear, 
if approximate, linear connection
between $1-\alpha$ and $\theta$.

Despite this linearity, we see that a pure Simon model fails
to accurately predict the phrase distribution exponent
$\theta$. 
This is not surprising, 
as when $\alpha \rightarrow 0$, 
an immediate adherence to the rich-get-richer
mechanism produces a transient behavior in which the first few 
(largest-count) word varieties exist out of proportion to
the eventual scaling. 
Because a pure Zipf/Simon distribution
preserves $\theta = 1 - \alpha$, 
we expect that a true, 
non-transient 
power-law consistently makes the underestimate
$1-N/M<\theta$.

\begin{figure*}[tp!]
  \centering
  \includegraphics[width=\textwidth]{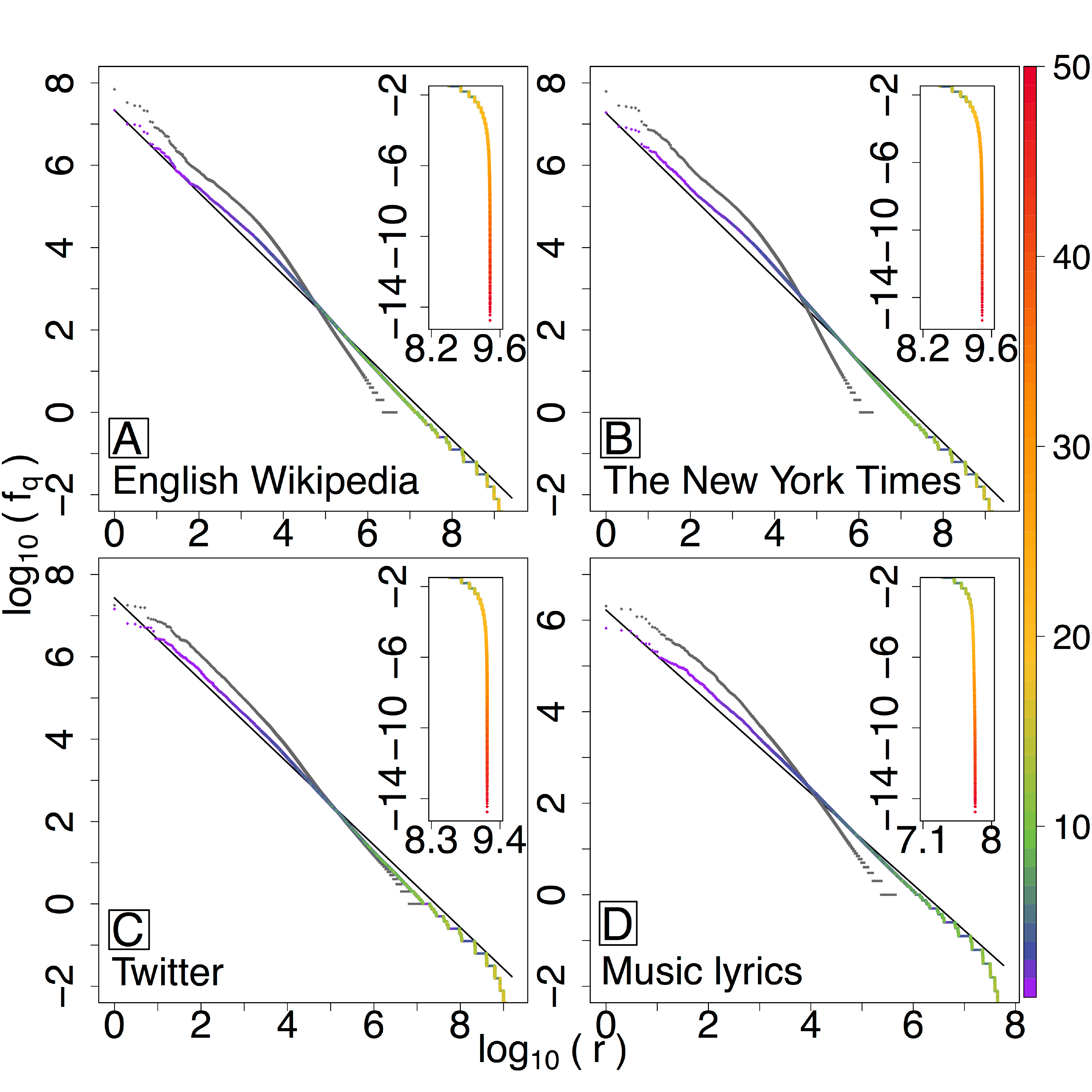}
  \caption{ 
    \protect
    Random partitioning distributions (\partitiontype{\onehalf}) for the four large
    corpora: 
    (A) Wikipedia (2010);
    (B) The New York Times (1987--2007); 
    (C) Twitter (2009); 
    and 
    (D) Music Lyrics (1960--2007).
    Top right insets show the long tails of random partitioning
    distributions, and the colors represent phrase length
    as indicated by the color bar.
    The gray curves are standard Zipf distributions
    for words (\partitiontype{1}), and exhibit limited scaling and
    with clear scaling breaks.
    See main text and 
    Tabs.~S1--S4,
    for example phrases.
  }
  \label{eq:tpz.randompartitionzipf}
\end{figure*}

Inspired by our results for one-off partitions of texts,
we now consider ensembles of pure random partitioning for larger texts.
In Fig.~\ref{eq:tpz.randompartitionzipf}, 
we show Zipf distributions of expected partition frequency, $f_q$,
for \partitiontype{\onehalf} phrases for four large-scale corpora: 
English Wikipedia, 
the New York Times (NYT),
Twitter,
and
music lyrics (ML),
coloring the main curves according to the length
of a phrase for each rank.
For comparison, 
we also include word-level Zipf distributions 
(\partitiontype{1})
for each text in gray,
along with the canonical Zipf 
distribution
(exponent $\theta$=1)
for reference.

We observe scalings for
the expected frequencies of phrases 
that hover around $\theta=1$
for over a remarkable 7--9 orders of magnitude.
We note that while others have observed similar results 
by simply combining frequency distributions of $N$-grams~\cite{ha2002a},
these approaches were unprincipled 
as they over-counted words.
For the randomly partitioned phrase distributions $f_\onehalf$, 
the scaling ranges we observe
persist down to $10^{-2}$, beyond 
the happax legomena,
which occur at frequencies greater
than $10^{-1}$.
Such robust scaling is in stark contrast to the very limited scaling 
of word frequencies (gray curves).
For pure word partitioning, \partitiontype{1}, 
we see two highly-distinct scaling regimes 
exhibited by each corpus, 
with shallow upper (Zipf) scalings at best 
extending over four orders of magnitude,
and typically only three.
(In a separate work, we investigate
this double scaling finding evidence that text-mixing
is the cause~\cite{williams2014b}.)

For all four corpora, random partitioning gives rise to 
a gradual interweaving of different length phrases when moving up through rank $r$.
Single words remain the most frequent (purple), typically 
beginning to blend with two word phrases (blue) by rank
$r=100$.  
After the appearance of phrases of length around
10--20, depending on the corpus,
we see the phrase rank distributions fall off sharply,
due to long clauses that are 
highly unique in their construction
(upper right insets).

In the supplementary material, we provide structured tables of
example phrases extracted by pure random partitioning for all four
corpora
(Tabs.~\ref{tab:tpz.wikipedia}--\ref{tab:tpz.lyrics}),
along with complete phrase data sets.
As with standard $N$-grams, the texture of each corpus is 
quickly revealed by examining phrases of length 3, 4, and 5.
For example, the second most common phrases of length 5 
for the four corpora are routinized phrases:
``the average household size was'' (EW),
``because of an editing error'' (NYT),
``i uploaded a youtube video'' (TW),
and
``na na na na na'' (ML).
By design, random partitioning allows us to 
quantitatively compare and sort phrases of different lengths.
For music lyrics, ``la la la la la'' has an expected frequency
similar to ``i don't know why'', ``just want to'', ``we'll have'',
and ``whatchu'' 
(see Tab.~\ref{tab:tpz.lyrics}),
while for the New York Times,
``the new york stock exchange'' is comparable
to ``believed to have'' 
(see Tab.~\ref{tab:tpz.nyt}).

\section{Discussion}

The phrases and their effective frequencies 
produced by our pure random partitioning method may serve 
as input to a range of higher order analyses.
For example, 
information theoretic work may be readily carried out,
context models may be built around phrase adjacency using
insertion and deletion,
and
specific, sentence-level partitions may be realized from probabilistic
partitions.

While we expect that other principled, more sophisticated approaches to partitioning
texts into rankable mixed phrases should produce Zipf's law spanning
similar or more orders of magnitude in rank,
we believe random partitioning---through its transparency, simplicity, and
scalability---will prove to be a powerful method for exploring
and understanding large-scale texts.

To conclude, our results reaffirm Zipf's law for language,
uncovering its applicability to a vast lexicon of phrases.
Furthermore, we demonstrate that the general semantic units 
of statistical linguistic analysis can and 
must be phrases---not words---calling for a reevaluation
and reinterpretation of past and present word-based studies
in this new light.

\acknowledgments
The authors are grateful for the computational resources provided by
the Vermont Advanced Computing Core which was supported by NASA (NNX
08A096G).
CMD was supported by NSF grant DMS-0940271;
PSD was supported by NSF CAREER Award \#0846668.

\clearpage

\newwrite\tempfile
\immediate\openout\tempfile=startsupp.txt
\immediate\write\tempfile{\thepage}
\immediate\closeout\tempfile

\setcounter{page}{1}
\renewcommand{\thepage}{S\arabic{page}}
\renewcommand{\thefigure}{S\arabic{figure}}
\renewcommand{\thetable}{S\arabic{table}}
\setcounter{figure}{0}
\setcounter{table}{0}

\section*{SI-1: Materials and methods}
To obtain the results in Fig.~\ref{fig:tpz.twocities_more},
we utilize the maximum likelihood estimation (MLE) procedure 
developed in~\cite{clauset2009b}.
In applying this procedure to clause and phrases distributions,
several quantities are generally considered:
\begin{itemize}
\item 
  $\hat{\theta}$: Zipf exponent estimate.
\item 
  $\xmin$: upper cutoff in rank $r$ determined by MLE procedure.
\item   
  $\kstat$: Kolmogorov-Smirnov (KS) statistic.
\item   
  $\clp$ determined by the MLE procedure (note that higher is better
  in that the null hypothesis is more favored).
\item   
  $1-\alpha$: Estimate of Zipf exponent $\theta$ based on
  Simon's model~\cite{simon1955a} where $\alpha$ is the
  introduction rate of new terms.
  We estimate $\alpha$ as the number of unique terms 
  ($N$) divided by the total number of terms ($M$).
\end{itemize}
which we report for 14 famous works of literature in SI-3.

In Fig.~\ref{fig:tpz.twocities_more}C we measure 
covariation between regressed values of $\hat{\theta}$ and
the Simon model prediction $1-\alpha$. 
Since both are subject to measurement error
($\hat{\theta}$ is a regressed quantity and 
$\alpha$ is only coarsely approximated by $N/M$),
we adhere to Reduced Major Axis regression \cite{rayner1985a},
which produces equivalent results upon interchanging
$x$ and $y$ variables,
and hence guarantees that no information is assumed or lost 
when we place $\hat{\theta}$ as the $x$-variable).

To produce the rank-frequency distributions
in Fig.~\ref{eq:tpz.randompartitionzipf}
and words in tables S1--S4,
we apply the random partition process
to several large corpora from
a wide scope of content.
These corpora are:
twenty years of New York Times articles (NYT, 1987--2007)~\cite{times2008},
approximately $4\%$ of a year's tweets (Twitter, 2009)~\cite{twitter2009},
music lyrics from thousands of songs and authors (Lyrics, 1960--2007)~\cite{dodds2009b},
and a collection of complete Wikipedia articles (Wikipedia, 2010)~\cite{wikipedia2010}.
In Fig.~\ref{fig:tpz.twocities_more} we also use a subset of more than $4,000$ books 
from the Project Gutenberg eBooks collection (eBooks, 2012)~\cite{gutenberg2010} 
of public-domain texts.

\section*{SI-2: Proof of $f_q$ word conservation}

In the body of this document we claim that
the random partition frequencies of the
phrases within a text $T$ conserve the text's
underlying mass of words, $M_T$.
This claim relies on the fact that the
partition frequencies of phrase-segments, $t_{i\cdots j}$, 
emerging from a single clause, $t$,
preserve its word mass, $\ell(t)$.
We represented this by the summation presented 
(Eq.~\ref{eq:tpz.phraseweight_text})
in the body of this document, 
which is equivalent to,
$f_q(S\mid t)E_S[\ell(s)\mid t]$,
i.e., the total number of words
represented by the frequency of appearance
of all phrases generated by the $q$-partition:
\begin{equation}
  \begin{split}
    f(S\mid t)E_S[\ell(s)\mid t]~&=~\sum_{s\in S}\ell(s)f_q(s\mid t)\\    
    &=~\sum_{s\in S}\sum_{s=t_{i\cdots j}}\ell(t_{i\cdots j})P_q(t_{i\cdots j}\mid t)\\
    &=~\sum_{1\leq i<j\leq\ell(t)}\ell(t_{i\cdots j})P_q(t_{i\cdots j} \,|\, t),
  \end{split}
\end{equation}
which we now denote by $M(S\mid t)$ for brevity.
For convenience, we now let $n=\ell(t)$ denote the clause's length
and observe that for each phrase-length $k<n$
there are two single-boundary phrases having partition probability
$q(1-q)^{k-1}$,
and $n-k-1$
no-boundary phrases having partition probability
$q^2(1-q)^{k-1}$.
The contribution to the above sum 
by all $k$-length phrases
is then given by 
\begin{equation}
  2kq(1-q)^{k-1} + (n-k-1)kq^2(1-q)^{k-1}.
\end{equation}
Upon noting the frequency 
of the single phrase 
(equal to the clause $t$)
whose length is $n$, $(1-q)^{n-1}$,
we consider the sum over all $k\leq n$,
\begin{equation}
\begin{split}
  M(S\mid t)~=~(1-q)^{n-1}&\\
  +~\:[2q+nq^2]\:&\sum_{k=1}^{n-1}k(1-q)^{k-1}\\
  -~q^2&\sum_{k=1}^{n-1}k(k+1)(1-q)^{k-1},
\end{split}
\end{equation}
which we will show equals $n$.
We now define the quantity $x=1-q$
(the probability that a space remains intact),
and in these terms find the sum to be:
\begin{equation}
\begin{split}
  M(S\mid t)~=~nx^{n-1}\hspace{69pt}&\\
   +~\left[2(1-x)+ n(1-x)^2\right]&\sum_{k=1}^{n-1}kx^{k-1}\\
   -~(1-x)^2&\sum_{k=1}^{n-1}k(k+1)x^{k-1}.
\end{split}
\end{equation}
This framing through $x$ affords a
nice representation in terms of the
generating function
\begin{equation}
  f(x)=\frac{1-x^{n+1}}{1-x},
\end{equation}
which allows us to express the summations through
derivatives of $f(x)$:
\begin{equation}
  \begin{split}
    &\sum_{k=1}^{n-1}kx^{k-1}=f'(x)-nx^{n-1}\text{, and}\\
    &\sum_{k=1}^{n-1}k(k+1)x^{k-1}=f''(x),
  \end{split}
\end{equation}
to find
\begin{equation}
  \begin{split}
    M(S\mid t)~&=~nx^{n-1}\\
    &+~\left[2(1-x)+n(1-x)^2\right](f'(x)-nx^{n-1})\\
    &-~(1-x)^2f''(x).
  \end{split}
\end{equation}
Substitution of the second derivative term
\begin{equation}
  f''(x)(1-x)=2f'(x)-n(n+1)x^{n-1}
\end{equation}
then produces the reduced form:
\begin{equation}
  \begin{split}
    M(S\mid t)~=~n&[f'(x)(1-x)^2\\
    &-~(nx^{n+1}-(n+1)x^n)],
  \end{split}
\end{equation}
into which we substitute the first derivative term
\begin{equation}
  f'(x)(1-x)^2=1+nx^{n+1}-(n+1)x^n,
\end{equation}
to render
\begin{equation}
  \begin{split}
    M(S\mid t)~=~n[1+&nx^{n+1}-(n+1)x^n\\
    -~(&nx^{n+1}-(n+1)x^n)]~=~n,
  \end{split}
\end{equation}
which proves Eq.~\ref{eq:tpz.phraseweight_text}.
Putting this together
into a sum over all clauses,
we see proof of Eq.~\ref{eq:tpz.wordmass_text} naturally follows:
\begin{equation}
  \begin{split}
    \sum_{s\in S}\ell(s)f_q(s\mid T)~&=~\sum_{t\in T}\sum_{s\in S}\ell(s)f_q(s\mid t)\\
    &=~\sum_{t\in T}M(S\mid t)~=~\sum_{t\in T}\ell(t).
  \end{split}
\end{equation}

\section*{SI-3: Parameters for well-known texts}
Below are tables showing fits of Zipf's exponent, $\hat{\theta}$,
for 14 famous works of literature,
along with details of the 
maximum likelihood estimation (MLE) procedure in~\cite{clauset2009b}.
The quantities used in these table are
described in SI-1, Materials and Methods.

\subsection*{A Tale of Two Cities}

\begin{tabular}{|c|c|c|c|c|c|}
\hline
 level & $\hat{\theta}$ & $\xmin$ & $\kstat$ & $\clp$ & $1-\alpha$ \\\hline
 clause & 0.783 & 3 & 0.0124 & 0.961 & 0.176 \\\hline
 phrase & 0.951 & 3 & 0.00742 & 0.772 & 0.603 \\\hline
 word & 1.15 & 4 & 0.0077 & 0.811 & 0.925 \\\hline
 grapheme & 1.56 & 4 & 0.0146 & 0.359 & 0.986 \\\hline
\end{tabular}

\subsection*{Moby Dick}

\begin{tabular}{|c|c|c|c|c|c|}
\hline
 level & $\hat{\theta}$ & $\xmin$ & $\kstat$ & $\clp$ & $1-\alpha$ \\\hline
 clause & 0.296 & 1 & 0.0192 & 0 & 0.154 \\\hline
 phrase & 0.902 & 3 & 0.0132 & 0.0626 & 0.576 \\\hline
 word & 1.05 & 7 & 0.00986 & 0.61 & 0.912 \\\hline
 grapheme & 1.42 & 13 & 0.0109 & 0.953 & 0.986 \\\hline
\end{tabular}

\subsection*{Great Expectations}

\begin{tabular}{|c|c|c|c|c|c|}
\hline
 level & $\hat{\theta}$ & $\xmin$ & $\kstat$ & $\clp$ & $1-\alpha$ \\\hline
 clause & 0.301 & 1 & 0.0199 & 0 & 0.186 \\\hline
 phrase & 0.995 & 5 & 0.0164 & 0.225 & 0.622 \\\hline
 word & 1.21 & 4 & 0.00943 & 0.526 & 0.938 \\\hline
 grapheme & 1.66 & 3 & 0.0147 & 0.181 & 0.988 \\\hline
\end{tabular}

\subsection*{Pride and Prejudice}

\begin{tabular}{|c|c|c|c|c|c|}
\hline
 level & $\hat{\theta}$ & $\xmin$ & $\kstat$ & $\clp$ & $1-\alpha$ \\\hline
 clause & 1 & 3 & 0.0204 & 0.911 & 0.172 \\\hline
 phrase & 0.983 & 3 & 0.0148 & 0.149 & 0.617 \\\hline
 word & 1.11 & 18 & 0.0201 & 0.662 & 0.947 \\\hline
 grapheme & 1.43 & 24 & 0.0226 & 0.698 & 0.989 \\\hline
\end{tabular}

\subsection*{Adventures of Huckleberry Finn}

\begin{tabular}{|c|c|c|c|c|c|}
\hline
 level & $\hat{\theta}$ & $\xmin$ & $\kstat$ & $\clp$ & $1-\alpha$ \\\hline
 clause & 0.881 & 4 & 0.0192 & 0.977 & 0.197 \\\hline
 phrase & 0.98 & 3 & 0.0119 & 0.385 & 0.625 \\\hline
 word & 1.47 & 1 & 0.0183 & 0.83 & 0.94 \\\hline
 grapheme & 1.66 & 6 & 0.0239 & 0.203 & 0.987 \\\hline
\end{tabular}

\subsection*{Alice's Adventures in Wonderland}

\begin{tabular}{|c|c|c|c|c|c|}
\hline
 level & $\hat{\theta}$ & $\xmin$ & $\kstat$ & $\clp$ & $1-\alpha$ \\\hline
 clause & 0.707 & 2 & 0.0198 & 0.711 & 0.191 \\\hline
 phrase & 0.906 & 2 & 0.0108 & 0.687 & 0.555 \\\hline
 word & 1.14 & 6 & 0.0353 & 0.105 & 0.899 \\\hline
 grapheme & 1.19 & 49 & 0.0338 & 0.972 & 0.975 \\\hline
\end{tabular}

\subsection*{The Adventures of Tom Sawyer}

\begin{tabular}{|c|c|c|c|c|c|}
\hline
 level & $\hat{\theta}$ & $\xmin$ & $\kstat$ & $\clp$ & $1-\alpha$ \\\hline
 clause & 0.321 & 1 & 0.0208 & 0 & 0.188 \\\hline
 phrase & 1.01 & 6 & 0.0173 & 0.826 & 0.555 \\\hline
 word & 1.12 & 3 & 0.0162 & 0.108 & 0.893 \\\hline
 grapheme & 1.51 & 4 & 0.0134 & 0.683 & 0.978 \\\hline
\end{tabular}

\subsection*{The Adventures of Sherlock Holmes}

\begin{tabular}{|c|c|c|c|c|c|}
\hline
 level & $\hat{\theta}$ & $\xmin$ & $\kstat$ & $\clp$ & $1-\alpha$ \\\hline
 clause & 0.308 & 1 & 0.0231 & 0 & 0.191 \\\hline
 phrase & 0.952 & 4 & 0.0093 & 0.892 & 0.586 \\\hline
 word & 1.09 & 9 & 0.0144 & 0.733 & 0.921 \\\hline
 grapheme & 1.44 & 12 & 0.0191 & 0.663 & 0.983 \\\hline
\end{tabular}

\subsection*{Leaves of Grass}

\begin{tabular}{|c|c|c|c|c|c|}
\hline
 level & $\hat{\theta}$ & $\xmin$ & $\kstat$ & $\clp$ & $1-\alpha$ \\\hline
 clause & 0.486 & 2 & 0.00768 & 0.783 & 0.0717 \\\hline
 phrase & 0.865 & 3 & 0.00971 & 0.463 & 0.543 \\\hline
 word & 1.01 & 6 & 0.0095 & 0.78 & 0.886 \\\hline
 grapheme & 1.39 & 7 & 0.0131 & 0.692 & 0.981 \\\hline
\end{tabular}

\subsection*{Ulysses}

\begin{tabular}{|c|c|c|c|c|c|}
\hline
 level & $\hat{\theta}$ & $\xmin$ & $\kstat$ & $\clp$ & $1-\alpha$ \\\hline
 clause & 0.34 & 1 & 0.0192 & 0 & 0.193 \\\hline
 phrase & 0.912 & 4 & 0.0062 & 0.854 & 0.551 \\\hline
 word & 1.05 & 5 & 0.00773 & 0.515 & 0.887 \\\hline
 grapheme & 1.48 & 4 & 0.00874 & 0.61 & 0.983 \\\hline
\end{tabular}

\subsection*{Frankenstein; Or, The Modern Prometheus}

\begin{tabular}{|c|c|c|c|c|c|}
\hline
 level & $\hat{\theta}$ & $\xmin$ & $\kstat$ & $\clp$ & $1-\alpha$ \\\hline
 clause & 0.257 & 1 & 0.0121 & 0 & 0.0741 \\\hline
 phrase & 0.834 & 2 & 0.0085 & 0.55 & 0.532 \\\hline
 word & 1.04 & 5 & 0.0215 & 0.057 & 0.906 \\\hline
 grapheme & 1.31 & 12 & 0.019 & 0.682 & 0.982 \\\hline
\end{tabular}

\subsection*{Wuthering Heights}

\begin{tabular}{|c|c|c|c|c|c|}
\hline
 level & $\hat{\theta}$ & $\xmin$ & $\kstat$ & $\clp$ & $1-\alpha$ \\\hline
 clause & 0.927 & 3 & 0.0217 & 0.751 & 0.178 \\\hline
 phrase & 0.952 & 7 & 0.0104 & 0.978 & 0.581 \\\hline
 word & 1.06 & 10 & 0.0163 & 0.533 & 0.917 \\\hline
 grapheme & 1.54 & 5 & 0.0165 & 0.345 & 0.984 \\\hline
\end{tabular}

\subsection*{Sense and Sensibility}

\begin{tabular}{|c|c|c|c|c|c|}
\hline
 level & $\hat{\theta}$ & $\xmin$ & $\kstat$ & $\clp$ & $1-\alpha$ \\\hline
 clause & 0.274 & 1 & 0.0176 & 0 & 0.142 \\\hline
 phrase & 0.982 & 3 & 0.00945 & 0.611 & 0.614 \\\hline
 word & 1.12 & 20 & 0.017 & 0.907 & 0.946 \\\hline
 grapheme & 1.41 & 28 & 0.0264 & 0.584 & 0.989 \\\hline
\end{tabular}

\subsection*{Oliver Twist}

\begin{tabular}{|c|c|c|c|c|c|}
\hline
 level & $\hat{\theta}$ & $\xmin$ & $\kstat$ & $\clp$ & $1-\alpha$ \\\hline
 clause & 0.93 & 3 & 0.0152 & 0.808 & 0.242 \\\hline
 phrase & 0.962 & 3 & 0.00945 & 0.439 & 0.622 \\\hline
 word & 1.13 & 8 & 0.0118 & 0.695 & 0.931 \\\hline
 grapheme & 1.52 & 7 & 0.0153 & 0.521 & 0.987 \\\hline
\end{tabular}

\clearpage

\section*{SI-4: Phrase frequency tables}

The following tables contain 
selected phrases extracted by random partitioning
for the four corpora examined in the main text.
We provide complete phrase lists in csv format along
with other material online at:
\href{http://www.uvm.edu/storylab/share/papers/williams2014a/}{http://www.uvm.edu/storylab/share/papers/williams2014a/}.

\begin{table*}[t]
  \begin{center}
\resizebox{!}{0.5\textheight}{
{\fontsize{7}{9}\selectfont
\begin{tabular}{|c|c|c|c|c|c|}
\hline rank & order=1 & order=2 & order=3 & order=4 & order=5 \\\hline
1&the (21763834.00)&of the (1332433.25)&one of the (42955.88)&in the united states (8425.91)&years of age or older (3363.23)\\\hline
2&in (9935182.25)&in the (1095178.50)&as well as (41878.69)&at the age of (5873.75)&the average household size was (1669.62)\\\hline
3&and (9708982.00)&to the (443282.25)&the united states (37460.25)&a member of the (5534.50)&were married couples living together (1662.12)\\\hline
4&of (9015261.00)&and the (404687.00)&part of the (23948.69)&under the age of (5287.88)&from two or more races (1530.73)\\\hline
5&a (6458405.25)&on the (335456.00)&at the time (17591.44)&the end of the (5013.12)&at the end of the (1512.25)\\\hline
6&to (5890435.75)&at the (308288.50)&the age of (17212.81)&at the end of (4780.31)&the median income for a (1251.14)\\\hline
7&was (3290575.00)&for the (282949.75)&the end of (16135.31)&as well as the (3805.84)&the result of the debate (1123.98)\\\hline
8&is (3203926.00)&he was (276889.75)&according to the (16111.19)&at the same time (3609.44)&of it is land and (863.06)\\\hline
9&he (2583977.75)&it is (246804.50)&may refer to (15914.88)&years of age or (3375.91)&the racial makeup of the (854.42)\\\hline
10&on (2577531.25)&with the (233894.38)&member of the (15805.50)&of age or older (3364.88)&has a total area of (847.59)\\\hline
11&as (2520721.50)&as a (230830.62)&the university of (15243.00)&the population density was (3354.00)&the per capita income for (841.80)\\\hline
12&for (2409743.75)&it was (209433.25)&a number of (14994.00)&the median age was (3332.41)&and the average family size (838.66)\\\hline
13&with (2107098.50)&from the (202985.38)&in the early (14390.50)&as of the census (3325.94)&and the median income for (832.59)\\\hline
14&by (2010245.50)&the first (177129.12)&as a result (14356.69)&households out of which (3290.84)&the average family size was (831.62)\\\hline
15&it (1960890.50)&as the (172026.62)&a member of (13692.75)&one of the most (2952.97)&had a median income of (831.28)\\\hline
16&from (1688878.50)&was a (153285.75)&in the united (13589.25)&people per square mile (2875.00)&of all households were made (830.97)\\\hline
17&that (1616682.00)&in a (152800.25)&he was a (13201.88)&at the university of (2866.38)&at an average density of (830.95)\\\hline
18&s (1588172.00)&to be (142233.38)&of the population (13129.81)&was one of the (2728.66)&males had a median income (830.89)\\\hline
19&at (1574302.50)&one of (128960.50)&in order to (12507.44)&for the first time (2684.28)&housing units at an average (829.80)\\\hline
20&his (1461713.50)&during the (128190.62)&was born in (11809.50)&the result of the (2675.75)&made up of individuals and (829.12)\\\hline
21&this (1187743.00)&of a (126613.62)&end of the (11779.88)&has a population of (2658.84)&had children under the age (828.27)\\\hline
22&an (1121850.50)&with a (120564.38)&in the late (11641.56)&on the other hand (2654.81)&someone living alone who was (827.98)\\\hline
23&are (965128.75)&and a (117848.38)&also known as (11477.12)&as part of the (2650.53)&income for a family was (825.89)\\\hline
24&or (962634.50)&such as (116356.12)&in addition to (11229.75)&of those under age (2626.69)&had someone living alone who (825.33)\\\hline
25&were (894722.00)&united states (107440.38)&it is a (11059.50)&during world war ii (2600.66)&householder with no husband present (823.45)\\\hline
26&also (771224.25)&as well (105543.38)&world war ii (11018.56)&of the united states (2591.84)&had a female householder with (816.72)\\\hline
27&be (736999.75)&th century (102688.62)&such as the (10948.00)&the median income for (2504.88)&population was spread out with (813.75)\\\hline
28&has (711456.75)&was the (102566.25)&the result was (10670.12)&as a result of (2403.53)&this is a list of (784.14)\\\hline
29&after (699095.75)&that the (98832.00)&most of the (10051.62)&he was born in (2381.19)&were below the poverty line (761.86)\\\hline
30&however (689592.50)&and was (93389.38)&as part of (9636.19)&to the united states (2366.12)&the united states census bureau (743.83)\\\hline
31&who (678548.00)&there were (88907.25)&he was the (9630.62)&in new york city (2292.53)&of the population were below (735.98)\\\hline
32&they (674922.00)&after the (86291.12)&due to the (9612.06)&in the summer of (2204.88)&according to the united states (734.53)\\\hline
33&one (657238.50)&new york (84445.25)&some of the (9501.06)&at the time of (2114.16)&was a member of the (711.28)\\\hline
34&she (628094.25)&citation needed (83924.50)&at the end (9428.75)&the rest of the (2013.94)&result of the debate was (692.72)\\\hline
35&had (625329.50)&he is (82930.25)&a population of (9329.19)&in the united kingdom (1997.19)&hispanic or latino of any (671.59)\\\hline
36&first (581708.50)&there are (81538.25)&it is the (9142.31)&as well as a (1862.66)&at the time of the (643.62)\\\hline
37&their (565228.50)&and in (79643.25)&there is a (9116.25)&of the population were (1848.16)&he was a member of (640.23)\\\hline
38&there (563650.50)&part of (79108.00)&new york city (8808.00)&the result was delete (1845.00)&it has a population of (583.19)\\\hline
39&when (554108.75)&for a (78926.88)&years of age (8519.56)&as one of the (1833.75)&national register of historic places (571.44)\\\hline
40&new (542938.25)&to a (75288.88)&members of the (8365.31)&a total area of (1827.31)&it had a population of (562.61)\\\hline
41&i (541554.00)&the united (73217.62)&of the year (8232.00)&was a member of (1718.38)&this list is intended to (554.30)\\\hline
42&its (540743.50)&has been (72469.62)&the city of (8122.94)&below the poverty line (1715.28)&it does not imply that (553.88)\\\hline
43&may (501345.50)&according to (71437.38)&this is a (8075.19)&the racial makeup of (1708.22)&interest in adding the link (553.47)\\\hline
44&have (501019.00)&of his (71418.38)&was the first (7792.12)&the per capita income (1692.88)&below a full report on (552.31)\\\hline
45&th (499317.50)&for example (67477.38)&was one of (7664.56)&the average household size (1691.25)&or that the involved accounts (550.88)\\\hline
46&her (490852.25)&the new (66748.88)&in the world (7551.44)&married couples living together (1681.62)&accounts are spamming the link (550.88)\\\hline
47&years (472193.25)&and is (65799.75)&in the first (7480.19)&a median income of (1677.53)&the external link gets used (549.97)\\\hline
48&m (470841.75)&the same (65643.25)&in new york (7475.44)&and the average family (1677.31)&the village has a population (541.09)\\\hline
49&all (448565.75)&the city (64700.38)&the united kingdom (7361.56)&average household size was (1670.56)&village has a population of (538.23)\\\hline
50&other (443913.00)&this is (61094.62)&in the u (7321.44)&median income for a (1668.47)&of the debate was delete (470.31)\\\hline
\hline100&under (265543.75)&who was (39346.88)&at the same (5320.88)&in adding the link (1106.97)&does not imply that involved (275.44)\\\hline
\hline150&because (181436.25)&and he (30431.00)&in the town (4118.00)&for a household in (831.50)&household in the town was (172.14)\\\hline
\hline200&games (136824.00)&who had (26478.88)&the poverty line (3444.69)&at the start of (677.09)&like some other vfd subpages (128.88)\\\hline
\hline250&still (114752.25)&the american (22596.00)&the soviet union (3232.75)&is located on the (551.88)&the united states department of (103.78)\\\hline
\hline300&great (98223.75)&the uk (19042.25)&in the region (2778.00)&a municipality in the (491.28)&in the first round of (91.98)\\\hline
\hline350&court (84906.00)&in late (16349.38)&name of the (2500.62)&afds for this article (422.00)&an archive of the discussion (84.06)\\\hline
\hline400&further (75941.25)&and are (14785.12)&the government of (2289.19)&one of the oldest (383.66)&income for the village was (78.66)\\\hline
\hline450&does (68667.00)&size was (13455.25)&as early as (2140.56)&the origin of the (343.66)&township has a total area (71.88)\\\hline
\hline500&wrote (60868.25)&less than (12458.00)&according to a (1982.56)&in the development of (320.78)&can be seen in the (66.02)\\\hline
\hline600&hit (51657.00)&place in (10748.25)&also refer to (1758.62)&the new york city (280.81)&of the new york times (56.72)\\\hline
\hline700&ground (44973.75)&he took (9618.50)&of the show (1622.31)&he studied at the (252.50)&archive of an rfd nomination (49.69)\\\hline
\hline800&lower (39646.75)&the canadian (8583.50)&so that the (1498.31)&in the fourth quarter (226.56)&of the church of england (44.61)\\\hline
\hline900&fall (34990.25)&republic of (7820.88)&of the french (1378.50)&played college football at (210.66)&peerage of the united kingdom (41.45)\\\hline
\hline1000&ad (31756.75)&with other (7171.38)&united states navy (1246.62)&city has a total (197.91)&fis nordic world ski championships (39.11)\\\hline
\hline1500&garden (21653.75)&from which (5251.62)&throughout the world (916.56)&at the outbreak of (154.88)&it is found in europe (30.12)\\\hline
\hline2000&ben (16336.25)&the need (4203.12)&of the oldest (753.75)&also part of the (125.28)&as a result of their (24.70)\\\hline
\hline2500&band's (12613.25)&formed the (3456.62)&a small village (640.88)&it was performed in (107.12)&the first game of the (21.09)\\\hline
\hline3000&modified (10278.00)&of information (2951.50)&season in the (562.62)&was entered into the (93.94)&the second single from the (18.69)\\\hline
\hline3500&md (8490.25)&system is (2609.38)&the highest point (501.00)&as late as the (84.00)&he became chairman of the (16.84)\\\hline
\hline4000&mythology (7228.25)&water and (2340.88)&team for the (455.06)&he served with the (77.25)&the most important of these (15.38)\\\hline
\hline4500&joan (6291.25)&in singapore (2118.25)&to the french (416.31)&majority of the population (71.12)&in burgundy in north-central france (14.12)\\\hline
\hline5000&politically (5492.50)&cold war (1922.25)&has an area (386.50)&old at the time (65.78)&station went on the air (13.20)\\\hline
\hline6000&ignored (4359.25)&be given (1641.62)&national association of (334.69)&claimed to be the (57.66)&the population has grown at (11.66)\\\hline
\hline7000&lexington (3519.25)&and perhaps (1437.00)&for two weeks (297.25)&there may be a (51.59)&the university of new brunswick (10.50)\\\hline
\hline8000&blackburn (2945.50)&much larger (1281.50)&cities in the (268.31)&advanced to the final (46.97)&of the church of st (9.58)\\\hline
\hline9000&eighteenth (2495.75)&the ride (1156.88)&with the american (244.19)&used to determine the (42.91)&whom he had two children (8.86)\\\hline
\hline10000&validity (2140.00)&bit of (1053.75)&birth to a (224.81)&of the songs on (39.78)&the new york court of (8.23)\\\hline
\hline15000&topical (1181.00)&of alcohol (726.62)&the australian national (162.31)&continued to serve as (29.50)&over the next three seasons (6.17)\\\hline
\hline20000&timeslot (765.75)&history is (556.75)&queen elizabeth i (127.94)&he was also chairman (23.97)&unsuccessful candidate for election in (5.03)\\\hline
\hline25000&wheatley (545.50)&the peruvian (451.38)&and a friend (107.12)&and asks him to (20.25)&during the making of the (4.31)\\\hline
\hline30000&epithelial (411.50)&can change (376.50)&it difficult for (92.19)&from george washington university (17.69)&during the two world wars (3.78)\\\hline
\hline35000&awakes (324.25)&footage from (324.50)&the march of (81.12)&with this surname include (15.75)&was the eldest child of (3.38)\\\hline
\hline40000&ruck (262.75)&marion county (283.88)&over the area (72.56)&please give a reliable (14.25)&the immigration and nationality act (3.06)\\\hline
\hline45000&verbandsliga (218.50)&a sculpture (252.88)&his father worked (65.56)&so in order to (13.06)&in new york during the (2.83)\\\hline
\hline50000&imageshack (185.25)&break with (227.12)&fire at the (59.88)&departed new york on (12.06)&and his family lived in (2.61)\\\hline
\hline60000&partito (140.50)&injured his (188.62)&for economic development (51.19)&were only able to (10.50)&the permanent court of arbitration (2.30)\\\hline
\hline70000&akatsuki (110.50)&ships had (160.88)&on the sky (44.88)&as an actress in (9.34)&was first known as the (2.05)\\\hline
\hline80000&salley (89.75)&various characters (140.12)&stanford university and (39.94)&there is some overlap (8.41)&the northeast end of the (1.86)\\\hline
\hline90000&huila (74.75)&the sodium (123.88)&though in a (36.06)&during the following decade (7.69)&the australian national rugby league (1.70)\\\hline
\hline100000&leaven (63.75)&the condemned (110.88)&other two were (32.88)&closely linked with the (7.06)&already at the age of (1.58)\\\hline
\end{tabular}
}
}
\end{center}
  
  \caption{
    Example phrases for English Wikipedia
    extracted by random partitioning.
  }
  \label{tab:tpz.wikipedia}
\end{table*}

\begin{table*}[t]
  \begin{center}
\resizebox{!}{0.5\textheight}{
{\fontsize{10}{12}\selectfont
\begin{tabular}{|c|c|c|c|c|c|}
\hline rank & order=1 & order=2 & order=3 & order=4 & order=5 \\\hline
1&the (19034045.00)&of the (922676.50)&the united states (48226.25)&in the united states (7162.22)&at the end of the (599.03)\\\hline
2&a (8722183.25)&in the (778571.88)&one of the (34160.31)&at the same time (5127.59)&because of an editing error (556.81)\\\hline
3&and (8175499.25)&he said (506762.62)&in new york (32747.94)&for the first time (3893.78)&the new york stock exchange (514.61)\\\hline
4&of (7463223.50)&to the (321805.25)&the new york (19706.31)&the new york times (3282.12)&for the first time in (481.62)\\\hline
5&to (7094522.25)&and the (312622.62)&as well as (19019.81)&in new york city (3036.69)&he is survived by his (478.02)\\\hline
6&in (6553996.25)&for the (275765.75)&new york city (17266.12)&at the end of (2664.31)&is survived by his wife (454.94)\\\hline
7&that (3251408.00)&at the (266174.25)&a lot of (14997.94)&the end of the (2560.50)&an initial public offering of (400.08)\\\hline
8&for (2849787.25)&new york (234356.50)&some of the (12923.62)&a spokesman for the (2556.88)&by the end of the (391.30)\\\hline
9&he (2720690.75)&in a (228202.25)&part of the (12009.06)&at the university of (2224.84)&the end of the year (354.31)\\\hline
10&is (2668672.00)&to be (182396.25)&of new york (11626.38)&one of the most (2167.66)&the securities and exchange commission (340.56)\\\hline
11&it (2252598.00)&with the (180261.50)&president of the (10928.75)&of the united states (2105.25)&for the first time since (328.12)\\\hline
12&but (2134976.50)&that the (179624.88)&the end of (10895.50)&a member of the (2028.19)&for students and the elderly (298.50)\\\hline
13&on (2102270.50)&it is (171736.38)&there is a (10682.38)&the rest of the (1907.81)&beloved wife of the late (292.89)\\\hline
14&with (2090580.50)&from the (165015.00)&director of the (10320.38)&at the age of (1877.81)&he said in an interview (287.44)\\\hline
15&at (2042863.25)&of a (161459.62)&it was a (10318.81)&to the united states (1832.50)&the dow jones industrial average (276.14)\\\hline
16&as (1808659.75)&she said (160297.25)&as a result (10075.00)&in lieu of flowers (1794.28)&the executive director of the (270.16)\\\hline
17&i (1626505.00)&by the (159916.25)&according to the (10053.56)&executive director of the (1718.41)&tonight and tomorrow night at (253.62)\\\hline
18&by (1573509.50)&it was (159603.00)&in the last (9828.88)&the united states and (1653.31)&in the last two years (243.44)\\\hline
19&his (1418411.25)&as a (146938.88)&the white house (9593.25)&is one of the (1549.75)&in the new york times (240.67)\\\hline
20&from (1397015.25)&he was (146862.00)&in the united (9578.31)&of the new york (1541.53)&in the last few years (235.52)\\\hline
21&who (1317491.75)&is a (142374.75)&the university of (9083.88)&by the end of (1524.62)&in the united states and (229.91)\\\hline
22&an (1253617.50)&with a (135244.50)&there is no (9027.81)&as well as the (1447.84)&in the middle of the (228.61)\\\hline
23&are (1179629.75)&and a (126899.75)&it is a (8987.25)&the chairman of the (1339.56)&there are a lot of (222.73)\\\hline
24&they (1177411.75)&but the (120749.75)&the first time (8735.56)&he is survived by (1330.34)&at the university of california (222.31)\\\hline
25&not (1163949.50)&one of (118009.62)&in the first (8607.00)&the new york city (1322.84)&the federal bureau of investigation (221.33)\\\hline
26&be (1140990.25)&for a (113570.88)&a spokesman for (8528.75)&in a telephone interview (1289.75)&the museum of modern art (220.48)\\\hline
27&this (1017793.00)&the new (107764.88)&at the time (8300.88)&at a news conference (1162.12)&of the new york times (214.25)\\\hline
28&which (985107.50)&the first (105144.75)&out of the (8246.56)&in the new york (1153.72)&graduated from the university of (210.23)\\\hline
29&or (927178.00)&united states (103164.62)&in the past (8010.69)&for the most part (1147.06)&the food and drug administration (207.61)\\\hline
30&new (892914.75)&as the (100548.38)&to be a (7877.38)&a son of mr (1144.06)&but at the same time (201.62)\\\hline
31&had (865149.00)&is the (95388.62)&this is a (7856.44)&a spokeswoman for the (1103.06)&as a result of the (200.59)\\\hline
32&one (826293.50)&will be (94356.50)&for the first (7789.44)&as a result of (1066.22)&the metropolitan museum of art (200.20)\\\hline
33&about (820268.00)&to a (92111.75)&in an interview (7685.56)&a lot of people (1060.12)&the university of california at (193.88)\\\hline
34&she (799892.00)&the united (91259.75)&he said he (7576.50)&a few years ago (1047.81)&years old and lived in (193.58)\\\hline
35&s (796792.25)&there is (83281.62)&the number of (7551.12)&of new york city (1034.91)&for the new york times (189.27)\\\hline
36&we (781654.50)&th street (81072.25)&of the new (7016.19)&new york stock exchange (1024.41)&received a master's degree in (179.23)\\\hline
37&when (752716.25)&for example (74955.88)&the same time (6904.50)&at a time when (1023.19)&a memorial service will be (179.17)\\\hline
38&will (704428.00)&according to (70748.12)&it was the (6859.56)&the director of the (1007.72)&new york and new jersey (176.58)\\\hline
39&there (700976.25)&would be (70553.75)&it would be (6843.44)&survived by his wife (998.25)&president and chief executive of (175.53)\\\hline
40&their (699595.50)&of his (70529.62)&in the world (6814.81)&as part of the (986.62)&president and chief operating officer (172.33)\\\hline
41&p (687358.75)&this is (69945.38)&it is not (6789.88)&in the middle of (970.16)&at the time of the (167.44)\\\hline
42&were (676437.25)&there are (69653.25)&in recent years (6653.56)&and the united states (956.59)&the rest of the world (165.12)\\\hline
43&years (672249.00)&that he (69545.88)&in the early (6652.31)&from the university of (916.47)&th street and amsterdam avenue (164.03)\\\hline
44&would (664100.25)&he is (69104.00)&in addition to (6584.25)&i don't want to (901.09)&the end of the day (156.91)\\\hline
45&you (616708.00)&they are (68165.50)&the united nations (6541.31)&in addition to the (897.94)&the united states court of (155.62)\\\hline
46&its (611930.00)&years ago (66357.25)&at the same (6344.44)&the first time in (897.12)&for more than a decade (151.02)\\\hline
47&if (608648.75)&when the (65028.62)&but it is (6272.62)&in an effort to (888.00)&this film is rated r (149.75)\\\hline
48&her (571742.75)&in his (62736.00)&at the end (6264.12)&as well as a (883.31)&spoke on condition of anonymity (148.44)\\\hline
49&all (568749.50)&who is (62527.25)&i don't think (6247.25)&in the first half (883.22)&court of appeals for the (148.30)\\\hline
50&been (552982.75)&and mr (61636.88)&i don't know (6171.06)&president and chief executive (882.94)&in the last five years (147.31)\\\hline
\hline100&here (259618.25)&to have (44901.62)&executive director of (4271.94)&in the middle east (614.88)&he graduated from the university (111.53)\\\hline
\hline150&st (168117.75)&trying to (32876.25)&and chief executive (3384.31)&tens of thousands of (498.16)&the virus that causes aids (92.28)\\\hline
\hline200&information (133141.25)&kind of (26368.75)&not going to (2943.56)&the heart of the (425.56)&secretary of state george p (79.97)\\\hline
\hline250&young (108081.25)&where he (21971.38)&a long time (2503.81)&the first half of (387.38)&came to the united states (69.20)\\\hline
\hline300&enough (93902.75)&he did (19303.00)&vice president for (2282.38)&for a total of (347.19)&salt and pepper to taste (63.61)\\\hline
\hline350&county (79788.75)&to pay (17182.75)&declined to comment (2116.62)&time on the market (315.03)&the new york city opera (58.09)\\\hline
\hline400&tax (72699.25)&the west (15687.75)&would like to (1993.69)&salt and freshly ground (288.91)&in state supreme court in (53.38)\\\hline
\hline450&became (65631.00)&to come (14304.75)&to more than (1884.75)&the vast majority of (272.25)&who is in charge of (49.69)\\\hline
\hline500&doing (59774.25)&the soviet (13439.88)&to build a (1787.25)&he said it was (257.59)&at the university of wisconsin (47.52)\\\hline
\hline600&quarter (51948.25)&a more (11832.00)&would be the (1569.44)&new york city police (231.50)&the good news is that (42.91)\\\hline
\hline700&someone (44616.75)&in november (10500.62)&a part of (1436.62)&state supreme court in (211.69)&he is also survived by (39.03)\\\hline
\hline800&weekend (39540.00)&get a (9667.62)&in a new (1314.62)&they don't want to (194.69)&in the next five years (36.31)\\\hline
\hline900&plays (35724.50)&given the (8871.62)&but for the (1227.50)&the last several years (184.69)&that he not be identified (33.95)\\\hline
\hline1000&ask (32280.00)&to show (8172.12)&they would be (1141.31)&those of us who (174.66)&upper east side of manhattan (31.92)\\\hline
\hline1500&reduce (21437.25)&in late (5853.50)&who heads the (872.69)&of the same name (135.16)&i don't know what to (24.73)\\\hline
\hline2000&seventh (15906.00)&and up (4597.88)&ought to be (721.38)&will continue to be (111.94)&the democratic congressional campaign committee (20.64)\\\hline
\hline2500&expansion (12556.75)&why the (3791.38)&of the biggest (621.31)&of the iraq war (97.66)&in the second half and (17.94)\\\hline
\hline3000&importance (10172.50)&and get (3287.00)&believed to have (545.69)&in front of his (86.75)&a good place to start (16.02)\\\hline
\hline3500&andy (8297.75)&idea that (2869.38)&he has made (492.75)&it is unclear how (78.78)&of the foreign relations committee (14.52)\\\hline
\hline4000&assessment (7023.75)&due to (2576.38)&of the report (453.06)&original moldings and detail (72.62)&there's no question about it (13.31)\\\hline
\hline4500&rye (6046.50)&which may (2336.75)&which he was (417.50)&the second and third (67.31)&the book review last year (12.30)\\\hline
\hline5000&officiated (5247.00)&ceremony at (2147.50)&affected by the (387.06)&to a multiyear contract (62.81)&the first day of school (11.50)\\\hline
\hline6000&distinctive (4090.00)&while others (1826.88)&economist at the (340.38)&to pay more than (55.56)&we are unable to acknowledge (10.25)\\\hline
\hline7000&racist (3296.75)&day for (1604.50)&the number to (305.81)&trinity college in hartford (50.03)&it is a question of (9.31)\\\hline
\hline8000&cracked (2726.75)&long term (1428.25)&and i hope (278.56)&the results have been (45.59)&filed in state supreme court (8.56)\\\hline
\hline9000&shrine (2306.25)&three and (1294.75)&throughout the state (256.19)&in the last seven (41.94)&the company went public in (7.92)\\\hline
\hline10000&handel's (1978.75)&new generation (1181.38)&of the home (236.62)&that the police had (39.03)&if there is such a (7.41)\\\hline
\hline15000&forgo (1063.50)&states supreme (818.12)&there are fewer (175.75)&its way through the (29.50)&of economics at the university (5.64)\\\hline
\hline20000&fujitsu (666.75)&come at (627.62)&a room with (140.88)&the history of american (24.12)&that donations be made to (4.62)\\\hline
\hline25000&refrained (456.50)&north fork (508.38)&to explain to (118.50)&and mayor david n (20.56)&that the soviet union would (3.97)\\\hline
\hline30000&tree' (335.25)&to disarm (427.50)&going for it (102.44)&the best they can (18.12)&a former republican senator from (3.52)\\\hline
\hline35000&afrikaans (256.00)&close and (367.12)&out of character (90.56)&end zone for a (16.22)&the east and the west (3.17)\\\hline
\hline40000&rushers (201.75)&by louis (321.62)&a maze of (81.19)&it also plans to (14.75)&and does not want to (2.89)\\\hline
\hline45000&andrews's (162.25)&after hitting (285.50)&sit in a (73.88)&the new law will (13.50)&he said the white house (2.67)\\\hline
\hline50000&hearne (133.00)&candidate is (256.88)&eastern european countries (67.81)&confirmed that he had (12.53)&it was the first victory (2.48)\\\hline
\hline60000&inxs (94.50)&accounting standards (213.50)&to use for (58.38)&of people in this (11.00)&and this was one of (2.19)\\\hline
\hline70000&airships (69.75)&compensation and (181.75)&doing enough to (51.31)&as if it could (9.81)&until the end of world (1.97)\\\hline
\hline80000&wei-sender (53.75)&dairy farmers (157.50)&he had missed (45.88)&new jersey attorney general (8.88)&game in the eighth inning (1.78)\\\hline
\hline90000&willan's (42.75)&table tennis (138.88)&and special events (41.50)&this is a town (8.12)&pleaded not guilty to all (1.64)\\\hline
\hline100000&prosecutable (35.00)&caught with (124.00)&you are ready (37.88)&i can't say enough (7.50)&the end of the new (1.53)\\\hline
\end{tabular}
}
}
\end{center}
  
  \caption{
    Example phrases for the New York Times
    extracted by random partitioning.
  }
  \label{tab:tpz.nyt}
\end{table*}

\begin{table*}[t]
  \begin{center}
\resizebox{!}{0.5\textheight}{
{\fontsize{10}{12}\selectfont
\begin{tabular}{|c|c|c|c|c|c|}
\hline rank & order=1 & order=2 & order=3 & order=4 & order=5 \\\hline
1&http (14482019.75)&in the (196458.75)&new blog post (34056.56)&i just took the (5910.19)&i favorited a youtube video (1839.47)\\\hline
2&com (6428552.75)&i am (157031.25)&check it out (18386.69)&e meu resultado foi (5061.88)&i uploaded a youtube video (1453.28)\\\hline
3&i (6227774.25)&i just (141596.00)&i love you (15578.25)&other people at http (3254.81)&just joined a video chat (1185.88)\\\hline
4&ly (5320341.75)&of the (140377.62)&i just took (15341.56)&check this video out (3243.06)&fiddling with my blog post (917.62)\\\hline
5&the (5180235.50)&on the (137894.88)&live on http (14544.62)&just joined a video (2371.72)&joined a video chat with (813.86)\\\hline
6&bit (5140992.75)&i love (137768.62)&i want to (13955.88)&a day using http (2061.75)&i rated a youtube video (642.88)\\\hline
7&a (5044536.50)&i have (136816.38)&i need to (12812.56)&on my way to (2006.66)&i just voted for http (582.91)\\\hline
8&to (4183208.25)&going to (121491.12)&i have a (12131.88)&favorited a youtube video (1842.59)&this site just gave me (581.73)\\\hline
9&o (2747181.50)&i think (120492.75)&quiz and got (11955.56)&i favorited a youtube (1839.59)&add a \#twibbon to your (472.08)\\\hline
10&rt (2735865.00)&to the (105588.38)&thanks for the (11796.62)&free online adult dating (1659.81)&the best way to get (454.42)\\\hline
11&and (2671876.00)&to be (103771.38)&what about you (10897.31)&a video chat with (1628.81)&just changed my twitter background (444.41)\\\hline
12&tinyurl (2630837.25)&i was (92118.25)&i think i (10602.31)&uploaded a youtube video (1461.91)&a video chat at http (375.84)\\\hline
13&you (2594872.75)&if you (89098.88)&i have to (10443.56)&i uploaded a youtube (1453.50)&photos on facebook in the (356.89)\\\hline
14&is (2589257.50)&at the (85136.38)&how are you (9339.94)&video chat at http (1435.12)&check it out at http (351.31)\\\hline
15&in (2278977.00)&i know (81260.50)&looking forward to (9084.25)&what do you think (1435.09)&own video chat at http (341.72)\\\hline
16&it (2209243.75)&have a (81252.50)&acabo de completar (9008.31)&i am going to (1398.34)&s channel on youtube http (304.75)\\\hline
17&me (2049880.50)&to get (79410.75)&i love it (8357.44)&if you want to (1359.84)&and won in \#mobsterworld http (293.28)\\\hline
18&d (2017195.75)&this is (78757.50)&a youtube video (8342.50)&i wish i could (1356.75)&live stickam stream at http (289.97)\\\hline
19&my (1967677.25)&and i (78420.62)&to go to (8035.69)&just got back from (1344.84)&on facebook in the album (289.03)\\\hline
20&of (1925590.25)&but i (77363.75)&of the day (8032.19)&at the same time (1310.53)&added myself to the http (275.00)\\\hline
21&on (1888067.25)&are you (76166.25)&what'll you get (7927.12)&thanks for the rt (1302.22)&just added myself to the (274.16)\\\hline
22&bitly (1814009.75)&it is (73377.25)&my daily twittascope (7900.69)&channel on youtube http (1284.31)&alot of followers using http (251.34)\\\hline
23&s (1644797.75)&i need (71952.50)&if you want (7526.38)&have a great day (1262.06)&has just done a job (235.92)\\\hline
24&n (1528713.00)&it was (70856.62)&going to be (7514.31)&joined a video chat (1186.28)&of followers check out http (232.44)\\\hline
25&lol (1514066.50)&is a (68456.88)&i don't know (7512.19)&is going to be (1167.62)&i love you so much (228.77)\\\hline
26&p (1439044.75)&i want (67746.75)&i wish i (7496.06)&trying to figure out (1145.16)&if you want to get (225.47)\\\hline
27&that (1325029.75)&i don't (67579.62)&is going to (7426.94)&thanks for the follow (1090.50)&hey i just got alot (202.48)\\\hline
28&at (1297915.75)&i can (67569.12)&going to bed (7393.75)&to your avatar now (1077.56)&mb and the humidity is (202.31)\\\hline
29&just (1260488.00)&to go (67275.00)&one of the (7351.12)&what are you doing (1069.03)&more followers go to http (201.09)\\\hline
30&u (1232849.00)&just voted (66498.25)&a lot of (7075.44)&can't wait to see (1031.53)&make your own video chat (200.94)\\\hline
31&de (1217523.25)&thank you (65912.62)&i feel like (7008.06)&com twitter directory under (1011.69)&you should check this site (199.98)\\\hline
32&lt (1189756.75)&want to (64412.75)&i just got (6909.56)&check it out http (1004.12)&site out if you want (197.48)\\\hline
33&this (1086481.00)&listening to (63914.62)&i need a (6752.81)&i have no idea (947.03)&where the wild things are (197.03)\\\hline
34&e (1076997.00)&in a (63833.88)&in the morning (6713.38)&add a \#twibbon to (944.44)&o \@luansantanaevc liga para voc (190.50)\\\hline
35&so (1069598.50)&right now (63056.50)&on my way (6635.31)&with my blog post (933.28)&joined a video chat at (186.12)\\\hline
36&www (1061481.75)&to do (61255.25)&let me know (6630.00)&i don't want to (929.91)&getting ready to go to (186.09)\\\hline
37&no (1042255.25)&have to (59581.12)&just took the (6078.25)&fiddling with my blog (924.09)&keep up the good work (183.22)\\\hline
38&gt (974932.50)&is the (58904.38)&meu resultado foi (6037.75)&i need to get (893.50)&gets you tons of followers (181.47)\\\hline
39&t (973331.25)&on my (58570.00)&can't wait to (5670.00)&i want to go (865.00)&i just become a member (178.52)\\\hline
40&with (964515.25)&you are (57962.88)&to be a (5662.25)&just got home from (861.56)&am i the only one (174.88)\\\hline
41&but (962587.00)&do you (57464.62)&just woke up (5596.50)&thank you so much (855.88)&let me know if you (174.84)\\\hline
42&im (961024.75)&at http (56957.25)&i just voted (5560.06)&the rest of the (851.44)&if you trying to get (173.91)\\\hline
43&now (942129.00)&i got (56806.25)&what do you (5274.19)&going to be a (848.88)&on my way to the (173.20)\\\hline
44&do (940366.75)&need to (56329.25)&just joined a (5162.19)&the best way to (831.09)&your own video chat at (170.86)\\\hline
45&m (935591.50)&vote too (56158.62)&i am so (5077.88)&i wish i was (817.91)&at the end of the (163.81)\\\hline
46&have (870991.75)&the best (55650.75)&e meu resultado (5062.62)&the end of the (789.47)&on my way to work (163.31)\\\hline
47&be (868701.00)&try it (52316.88)&in the world (4978.25)&check out this site (787.66)&trying to figure out how (163.19)\\\hline
48&twitpic (836558.50)&will be (51355.38)&happy new year (4964.62)&i can't wait to (784.03)&looking forward to your tweets (158.67)\\\hline
49&up (816280.25)&i will (50980.00)&getting ready to (4894.19)&i am listening to (760.44)&this is going to be (154.58)\\\hline
50&what (809227.00)&took the (50614.88)&getting ready for (4842.00)&to go to bed (757.81)&i want to go to (154.06)\\\hline
\hline100&know (434311.75)&on twitter (35430.38)&go to the (3525.19)&of followers using http (517.44)&sign up free and get (102.12)\\\hline
\hline150&come (265721.25)&good night (25500.38)&to see you (2833.94)&boa noite a todos (420.22)&don't know what to do (87.19)\\\hline
\hline200&watch (195026.50)&i mean (20696.62)&are you doing (2347.06)&more for gemini http (358.44)&the end of the world (65.86)\\\hline
\hline250&music (154807.00)&of it (18114.25)&at the end (2019.00)&a lot of people (310.47)&you know you want to (58.53)\\\hline
\hline300&soon (131072.75)&get the (15471.12)&out of my (1846.62)&all of a sudden (281.06)&us has given loan amount (52.56)\\\hline
\hline350&tell (113651.00)&i saw (13997.00)&to get the (1718.12)&i miss you too (259.38)&has been updated on nicedealz (48.59)\\\hline
\hline400&id (98939.25)&at a (12575.50)&to make it (1603.75)&a member of this (236.84)&now playing on smooth sounds (44.88)\\\hline
\hline450&gotta (85315.75)&i might (11760.38)&what happened to (1474.69)&how to make money (223.25)&i love him so much (42.11)\\\hline
\hline500&ne (76840.25)&able to (10855.50)&on the radio (1385.44)&calling it a night (209.53)&a day it work great (39.33)\\\hline
\hline600&care (63429.75)&my hair (9497.50)&to hear that (1256.88)&what do you want (191.09)&learn the trick discovered by (34.69)\\\hline
\hline700&once (52620.75)&as the (8335.25)&first day of (1133.31)&thanks for the \#followfriday (177.00)&i just snapped a new (31.94)\\\hline
\hline800&final (45764.00)&me out (7537.88)&day at work (1047.62)&going to bed early (159.59)&only a matter of time (29.42)\\\hline
\hline900&search (40726.00)&what u (6958.75)&all is well (957.31)&that would be a (148.81)&need to go to the (27.66)\\\hline
\hline1000&jackson (37068.25)&next year (6506.38)&be on the (906.00)&know what that means (139.56)&you want more followers check (26.00)\\\hline
\hline1500&program (24182.75)&but then (4861.50)&when you are (689.25)&is supposed to be (106.97)&just woke up from my (20.38)\\\hline
\hline2000&jones (17577.75)&reminds me (3908.38)&not the only (576.38)&to see you in (88.22)&made me laugh out loud (16.84)\\\hline
\hline2500&pengen (13684.50)&playing with (3222.12)&not sure i (495.19)&i feel for you (76.06)&my friend made this great (14.50)\\\hline
\hline3000&host (11228.75)&a person (2784.50)&they are the (435.19)&long way to go (67.12)&get in the way of (12.94)\\\hline
\hline3500&ghost (9481.00)&when a (2445.00)&there was no (386.44)&hope it gets better (60.88)&so sorry for your loss (11.73)\\\hline
\hline4000&chi (8112.25)&this a (2158.62)&and if i (350.12)&this minute was presented (55.94)&home for rent in houston (10.75)\\\hline
\hline4500&attempt (7138.25)&no homo (1952.12)&to do what (322.62)&i sound like a (52.03)&want to take a nap (9.97)\\\hline
\hline5000&strength (6254.00)&wanna get (1783.00)&i can just (298.00)&u know what i (48.66)&or however you spell it (9.33)\\\hline
\hline6000&andr (4908.00)&off your (1514.00)&don't know where (259.31)&best of luck to (43.06)&im going to take a (8.25)\\\hline
\hline7000&jeremy (3970.00)&a world (1312.62)&of things to (231.00)&will there be a (38.56)&medical and nursing staff wanted (7.44)\\\hline
\hline8000&domestic (3314.75)&em out (1177.38)&just voted demi (208.12)&what not to wear (35.31)&get more great followers at (6.81)\\\hline
\hline9000&aje (2837.00)&so wrong (1060.62)&yes you do (190.75)&you have a link (32.59)&it just me or are (6.31)\\\hline
\hline10000&fase (2468.50)&now following (966.38)&will be available (175.81)&to be in bed (30.25)&longs to be romantically entangled (5.91)\\\hline
\hline15000&lagunya (1421.00)&head off (662.12)&know each other (127.38)&will do my best (22.84)&it was good seeing you (4.47)\\\hline
\hline20000&bts (946.00)&lovin the (502.75)&is in need (101.44)&we should meet up (18.56)&a little bit of both (3.70)\\\hline
\hline25000&grandino (690.00)&more photos (407.75)&in the light (84.62)&join our site for (15.72)&going to see where the (3.19)\\\hline
\hline30000&treysongz (536.50)&jus had (342.38)&in this week (73.31)&a break from work (13.81)&this is a great site (2.80)\\\hline
\hline35000&helluva (438.00)&some guys (295.38)&too hot for (64.69)&would like to share (12.34)&i am so sorry i (2.50)\\\hline
\hline40000&woots (368.00)&very scary (259.12)&don't get mad (57.94)&have to stay in (11.16)&how bad can it be (2.28)\\\hline
\hline45000&combi (315.00)&red dwarf (230.75)&hard is it (52.69)&see some of you (10.22)&we are working on the (2.09)\\\hline
\hline50000&rfd (274.25)&recession proof (208.00)&little help from (48.25)&giving it a try (9.44)&to your sign today can (1.95)\\\hline
\hline60000&wwwstickamjp (217.00)&video converter (173.25)&i woulda been (41.50)&want to say i (8.22)&a long time ago i (1.72)\\\hline
\hline70000&casbah (180.00)&wonderful evening (148.12)&holidays to all (36.31)&vu all over again (7.31)&don't have enough on my (1.55)\\\hline
\hline80000&coolman (151.00)&via de (129.62)&min to get (32.38)&i thought of u (6.59)&but i can't take it (1.41)\\\hline
\hline90000&caius (123.25)&a share (115.00)&that's just sad (29.31)&lady gaga or beyonce (6.03)&my body wants to be (1.28)\\\hline
\hline100000&aliena (102.50)&class right (103.25)&eating a bowl (26.75)&i can't imagine that (5.56)&never do much of anything (1.19)\\\hline
\end{tabular}
}
}
\end{center}
  
  \caption{
    Example phrases for Twitter
    extracted by random partitioning.
  }
  \label{tab:tpz.twitter}
\end{table*}

\begin{table*}[t]
  \begin{center}
\resizebox{!}{0.5\textheight}{
{\fontsize{10}{12}\selectfont
\begin{tabular}{|c|c|c|c|c|c|}
\hline rank & order=1 & order=2 & order=3 & order=4 & order=5 \\\hline
1&i (668838.75)&in the (28174.25)&i love you (2556.75)&la la la la (514.06)&la la la la la (184.89)\\\hline
2&you (600813.50)&and i (25040.88)&i don't know (2094.00)&i don't want to (315.31)&na na na na na (93.98)\\\hline
3&the (576318.50)&i know (17993.00)&i want to (1750.06)&na na na na (281.78)&on and on and on (48.28)\\\hline
4&and (440698.25)&you know (16977.75)&la la la (1449.50)&in love with you (237.28)&i want you to know (47.70)\\\hline
5&to (330196.75)&i don't (16237.12)&i want you (1229.00)&i want you to (227.75)&you know what i mean (45.64)\\\hline
6&me (305085.75)&on the (14977.12)&you and me (1159.00)&i don't know what (201.38)&don't know what to do (45.22)\\\hline
7&a (301126.50)&if you (13856.62)&i don't want (1105.88)&i don't know why (187.59)&oh oh oh oh oh (40.80)\\\hline
8&it (219505.25)&to me (13048.50)&i know you (1086.00)&oh oh oh oh (181.59)&da da da da da (40.41)\\\hline
9&my (205611.00)&to the (12940.75)&i need you (1065.12)&i want to be (172.69)&do do do do do (40.02)\\\hline
10&in (203916.25)&to be (12614.00)&and i know (1051.62)&know what to do (144.06)&one more chance at love (35.66)\\\hline
11&that (150464.50)&i can (12372.12)&i don't wanna (914.00)&what can i do (141.41)&i don't want to be (35.38)\\\hline
12&of (149402.75)&and the (11679.88)&i got a (904.25)&yeah yeah yeah yeah (138.19)&in the middle of the (34.66)\\\hline
13&on (143576.50)&but i (11512.50)&i know that (903.00)&you don't have to (137.38)&i don't give a fuck (33.81)\\\hline
14&your (135024.00)&of the (11239.88)&you know i (902.69)&i close my eyes (130.31)&yeah yeah yeah yeah yeah (33.05)\\\hline
15&but (132235.00)&i can't (10372.88)&i can see (872.62)&you want me to (129.19)&i don't know what to (32.39)\\\hline
16&all (124985.50)&for you (10147.75)&and i don't (844.81)&you make me feel (128.31)&all i want is you (31.78)\\\hline
17&so (121375.75)&when i (10046.38)&in your eyes (844.06)&i just want to (128.00)&you know i love you (26.88)\\\hline
18&no (116877.00)&come on (9924.25)&i don't care (832.06)&da da da da (123.78)&the middle of the night (26.73)\\\hline
19&we (113865.25)&you can (9686.00)&and if you (825.94)&if you want to (123.06)&the rest of my life (26.34)\\\hline
20&is (113375.25)&i got (9577.88)&the way you (824.94)&come back to me (121.56)&no no no no no (26.11)\\\hline
21&for (108828.50)&in my (9473.12)&all the time (817.62)&in the middle of (119.16)&at the end of the (25.30)\\\hline
22&oh (107477.25)&all the (9467.25)&na na na (790.38)&and i don't know (118.72)&i wanna be with you (22.77)\\\hline
23&be (107432.75)&i want (9396.50)&don't you know (766.62)&let me tell you (117.66)&all i wanna do is (22.44)\\\hline
24&love (104438.50)&that i (9190.88)&this is the (766.25)&give it to me (111.97)&no matter what i do (22.41)\\\hline
25&it's (99026.75)&i am (9141.88)&can't you see (761.19)&you are the one (111.94)&the way you love me (21.42)\\\hline
26&now (95016.75)&and you (9048.75)&you love me (753.44)&do do do do (111.28)&no matter what you do (21.36)\\\hline
27&don't (94956.00)&i was (9028.12)&oh oh oh (749.56)&i love you so (111.16)&what you do to me (20.83)\\\hline
28&yeah (92807.00)&tell me (8783.50)&i wanna be (744.50)&all i want is (109.81)&when i close my eyes (20.31)\\\hline
29&when (91600.75)&like a (8614.12)&you know that (714.38)&how does it feel (109.69)&and i don't know why (20.09)\\\hline
30&with (90323.75)&the way (8512.38)&you want to (709.62)&know what i mean (109.12)&let me be the one (19.86)\\\hline
31&what (90190.50)&to you (8289.50)&you don't know (707.62)&no no no no (104.03)&the end of the day (18.64)\\\hline
32&this (90120.00)&when you (8157.62)&in my heart (693.69)&to be with you (100.81)&in the name of love (18.50)\\\hline
33&know (89600.00)&if i (7941.50)&you and i (691.50)&i don't wanna be (97.50)&lemme see you drip sweat (18.00)\\\hline
34&like (84259.00)&in a (7893.38)&you make me (675.19)&and on and on (96.47)&i like the way you (17.91)\\\hline
35&just (83346.75)&my heart (7882.88)&if you want (663.81)&the end of the (94.66)&it's been a long time (17.89)\\\hline
36&baby (83182.75)&for me (7880.50)&yeah yeah yeah (662.38)&i wish i could (93.09)&till the end of time (17.67)\\\hline
37&do (81926.00)&this is (7754.62)&don't want to (654.62)&don't give a fuck (92.94)&i wish that i could (17.61)\\\hline
38&up (81529.00)&for the (7570.88)&want to be (624.56)&can you feel it (91.88)&if you want me to (17.47)\\\hline
39&if (74941.25)&let me (7539.25)&in my life (622.44)&the way i feel (91.00)&see it in your eyes (17.20)\\\hline
40&chorus (72833.00)&with you (7482.62)&if i could (619.25)&i don't know how (90.47)&no matter what they say (16.78)\\\hline
41&can (67057.50)&i need (7424.62)&you know what (615.06)&gon play with it (90.00)&and i don't know what (16.73)\\\hline
42&down (66636.75)&with me (7386.00)&what you want (605.19)&you know that i (89.84)&let me hear you say (16.70)\\\hline
43&get (63408.50)&you are (7208.25)&i used to (604.88)&at the end of (89.38)&i look into your eyes (16.70)\\\hline
44&time (62579.50)&i wanna (7083.00)&on and on (595.94)&can you hear me (89.06)&i love the way you (16.64)\\\hline
45&out (62562.50)&what you (6949.00)&i see you (592.88)&want you to know (88.38)&and i don't want to (16.45)\\\hline
46&go (62101.75)&love you (6900.38)&in the sky (587.75)&out of my mind (86.62)&when i think of you (16.38)\\\hline
47&quot (61793.50)&the world (6774.62)&in the air (584.06)&i need to know (86.56)&i look in your eyes (16.31)\\\hline
48&got (60347.00)&do you (6733.50)&what to do (577.12)&all i wanna do (84.03)&the end of the world (16.16)\\\hline
49&one (59306.50)&from the (6679.88)&all night long (558.19)&on the other side (83.88)&when the sun goes down (16.11)\\\hline
50&see (58662.50)&want to (6649.88)&i know i (557.00)&do you love me (83.72)&still in love with you (16.02)\\\hline
\hline100&that's (28709.75)&in love (4324.38)&i just want (441.88)&that you love me (61.00)&you want me to do (11.83)\\\hline
\hline150&always (17981.50)&i won't (3225.00)&make me feel (369.31)&take a look at (51.09)&the end of the line (9.78)\\\hline
\hline200&en (13668.25)&without you (2692.50)&for you to (308.31)&you make me wanna (43.81)&that's the way it goes (8.77)\\\hline
\hline250&side (10606.00)&when i'm (2280.75)&who i am (277.81)&the rest of my (39.50)&the way that you do (7.88)\\\hline
\hline300&words (8896.00)&so long (2050.12)&on the wall (254.81)&open up your eyes (36.66)&i want to see you (7.23)\\\hline
\hline350&coming (7424.50)&have a (1815.25)&no one else (236.19)&get out of my (34.09)&makes the world go round (6.59)\\\hline
\hline400&ground (6669.25)&that's the (1645.12)&that's what i (218.94)&i don't want no (31.16)&tell me what you need (6.22)\\\hline
\hline450&death (5688.75)&then you (1506.12)&come back to (206.38)&don't mean a thing (29.25)&hey hey hey hey hey (5.81)\\\hline
\hline500&slow (5006.25)&i try (1382.25)&just want to (194.12)&goes on and on (27.84)&my my my my my (5.44)\\\hline
\hline600&cut (3808.00)&here i (1196.62)&i see your (172.44)&me like you do (25.44)&hey ladies drop it down (5.00)\\\hline
\hline700&grow (3091.25)&love with (1066.25)&in the game (158.62)&in front of you (23.47)&don't know if i can (4.61)\\\hline
\hline800&shut (2569.75)&my hands (969.25)&not the same (145.62)&you broke my heart (21.91)&it's been so long since (4.30)\\\hline
\hline900&doo (2167.75)&i tell (879.75)&yes i am (134.25)&me what you want (20.78)&you were the only one (4.06)\\\hline
\hline1000&seven (1898.75)&s a (802.88)&it was the (126.00)&all that i want (19.69)&just the way it is (3.88)\\\hline
\hline1500&food (1140.25)&am the (562.75)&a whole lot (95.38)&i wanna thank you (15.59)&mean a thing to me (3.20)\\\hline
\hline2000&fields (776.75)&caught up (434.12)&give me love (79.25)&got nothing to say (13.09)&a shoulder to cry on (2.78)\\\hline
\hline2500&vie (575.50)&saturday night (352.00)&yes you are (68.12)&know that you can (11.62)&was it good for you (2.48)\\\hline
\hline3000&compromise (451.00)&of things (295.38)&all about the (60.12)&is how we do (10.34)&right round like a record (2.27)\\\hline
\hline3500&couch (363.00)&the white (254.50)&think you can (53.75)&joy to the world (9.38)&your love would be untrue (2.08)\\\hline
\hline4000&pu (301.25)&they see (223.75)&i can fly (49.00)&if i don't get (8.69)&he was the only one (1.94)\\\hline
\hline4500&collect (254.75)&we'll have (197.62)&you said you'd (44.81)&give it all to (8.09)&you that we won't stop (1.81)\\\hline
\hline5000&product (219.25)&you drive (179.12)&want to hold (41.25)&wanna get with you (7.59)&cut me down to size (1.72)\\\hline
\hline6000&whatchu (169.50)&where you're (149.50)&take a breath (35.94)&your eyes on me (6.78)&round the ole oak tree (1.56)\\\hline
\hline7000&battered (135.25)&a plane (128.62)&right here in (32.00)&i wish i may (6.19)&move on down the line (1.44)\\\hline
\hline8000&verloren (111.25)&step out (111.88)&of all that (29.00)&we can make love (5.69)&bow wow wow yippie yo (1.33)\\\hline
\hline9000&nt (93.25)&fuck what (99.12)&be waiting for (26.44)&who the fuck are (5.25)&to warm a lonely night (1.25)\\\hline
\hline10000&honda (79.75)&you should've (88.38)&that what you (24.38)&like a loaded gun (4.88)&ain't that what you said (1.19)\\\hline
\hline15000&fuma (43.75)&little angel (57.50)&i wouldn't mind (17.44)&it's better this way (3.75)&on christmas day in the (0.94)\\\hline
\hline20000&cooper (28.50)&the undertow (42.00)&the wrong place (13.69)&since she left me (3.09)&and let it all go (0.80)\\\hline
\hline25000&fishy (20.25)&a major (32.88)&for one last (11.38)&and maybe you can (2.66)&no matter how far away (0.70)\\\hline
\hline30000&illtown (15.25)&alright baby (27.00)&you should try (9.75)&it take to make (2.34)&t want french fried potatoes (0.62)\\\hline
\hline35000&ndelo (12.00)&loud enough (22.62)&never give it (8.56)&i came to bring (2.12)&what you gave to me (0.58)\\\hline
\hline40000&rees (9.75)&view mirror (19.50)&to me a (7.62)&things i'm gonna do (1.94)&love is out the door (0.53)\\\hline
\hline45000&metaphoric (8.00)&your concern (17.12)&roll roll roll (6.88)&gotta say too much (1.75)&non ci sono solo io (0.50)\\\hline
\hline50000&memorizing (6.75)&the cancer (15.25)&on the eyes (6.25)&lay on the floor (1.62)&set the floor on fire (0.48)\\\hline
\hline60000&ajai (5.00)&an' then (12.38)&keep my eye (5.31)&give up on yourself (1.44)&right here next to you (0.44)\\\hline
\hline70000&aleiki (4.00)&cats be (10.38)&no more runnin' (4.62)&there's only one god (1.31)&gates of the seven seals (0.38)\\\hline
\hline80000&saatanan (3.25)&blijf ik (8.88)&we'll show them (4.12)&skies from now on (1.19)&we don't even have to (0.38)\\\hline
\hline90000&sauber (2.75)&yo tell (7.75)&time you say (3.69)&it comes to that (1.09)&ooh when you walk by (0.34)\\\hline
\hline100000&mosques (2.25)&believe anymore (6.88)&seemed so right (3.38)&but if i leave (1.00)&van de kille stemmen die (0.31)\\\hline
\end{tabular}
}
}
\end{center}
  
  \caption{
    Example phrases for Music Lyrics
    extracted by random partitioning.
  }
  \label{tab:tpz.lyrics}
\end{table*}

\end{document}